%% file: root.tex
\documentclass[letterpaper,10pt,conference]{ieeeconf}

\IEEEoverridecommandlockouts
\usepackage{cite}
\usepackage{amsmath,amssymb,bm}
\usepackage{booktabs,multirow,tabularx,array}
\usepackage[table]{xcolor}
\usepackage{xspace}
\usepackage{graphicx}
\graphicspath{{figures/}{GI_WM__Generalizable_Interaction_World_Modeling_with_Visual_Actions-4/}}
\usepackage{siunitx}
\usepackage{url}
\usepackage{dblfloatfix}
\usepackage{capt-of}
\usepackage[hidelinks]{hyperref}
\hypersetup{pdfauthor={Anonymous Authors}}

\definecolor{bestgreen}{RGB}{218,230,205}
\definecolor{secondgreen}{RGB}{240,245,233}
\newcommand{\bestcell}[1]{\cellcolor{bestgreen}\textbf{#1}}
\newcommand{\secondcell}[1]{\cellcolor{secondgreen}\underline{#1}}

\definecolor{webcolor}{HTML}{90A9B6}
\newcommand{\website}{\href{https://chenghaogu.github.io/GeniWorld/}{\textcolor{webcolor}{https://chenghaogu.github.io/GeniWorld/}}}

\title{\LARGE \bf
GeniWorld: A Generalizable Interactive World Model for \\
Robotic Manipulation via Visual Actions
}

\author{Chenghao Gu$^{1,2,*}$, Hanyang Yu$^{2,3,*}$, Jingbo Zhang$^{2,\dagger}$, Haitao Lin$^{2}$, Wenyao Zhang$^{2}$,\\
Jinghe Wang$^{1}$, Hanglei Jin$^{1}$, Shuzhao Xie$^{1}$, Jingyan Jiang$^{4}$, and Zhi Wang$^{1,\dagger}$\\[0.5em]
{\small $^{1}$Shenzhen International Graduate School, Tsinghua University \quad
$^{2}$Tencent Robotics X}\\
{\small $^{3}$The Hong Kong University of Science and Technology \quad
$^{4}$Shenzhen Technology University}\\
{\small $^{*}$Equal contribution \quad $^{\dagger}$Corresponding authors}\\
[0.4em]
{\small Project Page: \website}}

\IEEEaftertitletext{%
\begin{minipage}{\textwidth}
    \centering
    \vspace{-1.2em}
    \includegraphics[width=0.99\textwidth]{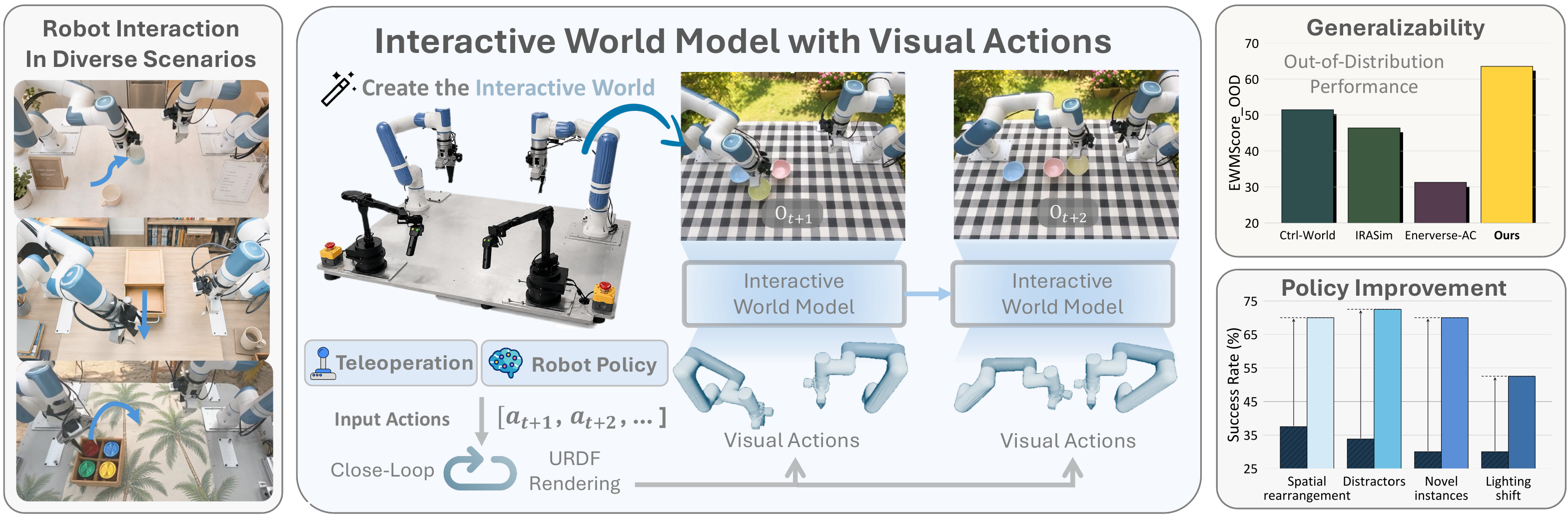}
    \captionof{figure}{\textbf{Overview of GeniWorld.} GeniWorld is an autoregressive robotic world model that transforms action inputs into visual action representations, enabling closed-loop interaction with human operators and robot policies. Trained on limited scene-specific demonstrations, GeniWorld generalizes to out-of-distribution (OOD) scenarios to produce high-fidelity observation predictions. Furthermore, it synthesizes rich manipulation trajectories to boost downstream policy performance and robustness under diverse conditions.}
    \label{fig:overview}
    \vspace{0.4em}
\end{minipage}%
}

\begin{document}

\maketitle
\thispagestyle{empty}
\pagestyle{empty}

%%%%%%%%%%%%%%%%%%%%%%%%%%%%%%%%%%%%%%%%%%%%%%%%%%%%%%%%%%%%%%%%%%%%%%%%%%%%%%%%
\begin{abstract}

Generalist robot policies exhibit strong capabilities, but their robustness in complex and unseen environments remains limited. Scaling robot learning and evaluation in diverse real-world environments remains costly and challenging. Action-conditioned world models offer a promising alternative, but they often suffer from limited action controllability and poor generalization to out-of-distribution (OOD) scenarios. To this end, we present \textbf{GeniWorld}, an interactive world model for robots that generalizes robustly across unseen scenarios. Building on pretrained video generative models, we use URDF-based rendering to transform numerical actions into visual action representations, enabling spatially grounded action control. By explicitly decoupling embodiment kinematics from environmental dynamics, our model mitigates scene overfitting and facilitates modeling of robot-environment interactions. To achieve closed-loop control, we construct an autoregressive video prediction model integrated with high-frequency robot kinematic control, enabling interaction with both robot policies and human teleoperators. In our experiments, even when trained solely on limited fixed-scene data, our model achieves superior in-domain performance and robust zero-shot generalization to highly randomized, unseen environments. For downstream applications, GeniWorld serves as a scalable policy evaluator that remains reliable under environmental perturbations. Furthermore, even with limited real-world demonstrations, GeniWorld generates diverse manipulation trajectories within the world model, improving downstream policy performance and robustness in complex environments.

\end{abstract}

\input{sections/1-intro.tex}
\input{sections/2-related.tex}
\input{sections/3-method.tex}

\input{sections/4-experiment.tex}
\input{sections/5-conclusion.tex}

\bibliographystyle{IEEEtran}
\bibliography{references}

\input{sections/X_appendix.tex}

\end{document}

%% file: sections/1-intro.tex
%%%%%%%%%%%%%%%%%%%%%%%%%%%%%%%%%%%%%%%%%%%%%%%%%%%%%%%%%%%%%%%%%%%%%%%%%%%%%%%%
\section{Introduction}
\label{sec:introduction}

Recent advances in generalist robot policies~\cite{black2024pi_0, intelligence2025pi, intelligence2025pi_, intelligence2026pi, bjorck2025gr00t, team2025gemini, lin2026posevla, zhang2026dreamvla, li2026causal, ye2026world, zhang2026disentangled, yuan2026fast} have demonstrated impressive multi-task capabilities. However, these policies often remain brittle when deployed in complex and unstructured environments~\cite{zhou2025libero, fei2025libero, chen2025robotwin}. Changes in backgrounds, object instances, spatial layouts, and task configurations can lead to substantial performance degradation~\cite{yuan2026qwen, yu2026maskwam}. Scaling robot learning and evaluation to cover such variations typically requires constructing new physical environments, which is costly and severely limits scene diversity.

Action-conditioned world models offer a promising alternative by enabling robots to interact within an imagination space, thereby facilitating scalable policy learning and evaluation. Existing approaches~\cite{guo2025ctrlworld, zhu2025irasim, li2025worldeval, quevedo2025worldgym, wang2026interactive} typically condition generative models directly on numerical action vectors. However, this paradigm suffers from two critical limitations. First, numerical action vectors lack explicit spatial grounding, limiting the accurate modeling of complex robot motions.
% Typically, existing approaches condition the generative model directly on numerical action vectors. However, they suffer from two critical limitations. First, numerical action representations lack precise spatial control over robot motion and struggle to model the expressive kinematic space.
Second, low-dimensional action conditioning entangles embodiment motion with environmental changes, forcing the model to focus excessively on interaction-irrelevant details and generalize poorly to unseen scenes.
% Specifically, we argue that conditioning the world model directly on robot action
We argue that directly conditioning the world model on robot motion provides precise spatial guidance for manipulation and captures fine-grained interaction details. Meanwhile, conditioning scene dynamics on robot motion more directly reflects the physical changes driven by robot--environment interactions.

To this end, we present \textbf{GeniWorld}, a generalizable interactive world model conditioned on embodied visual actions that enables closed-loop interaction with policies and human operators. Even with limited demonstrations, our model generalizes robustly across diverse scenarios and enables the generation of novel behaviors.

For the target robotic system, an embodiment-specific kinematic model first converts numerical action sequences into dense robot motion sequences. Building on a pretrained video generative model~\cite{wan2025wan}, GeniWorld encodes visual actions and scene observations into spatially aligned latent representations. We then build an autoregressive model with causal attention, ensuring that future predictions depend strictly on current robot actions and historical states. During inference, we leverage KV caching to maintain high-quality video generation while enabling closed-loop interaction.

We conduct comprehensive experiments to evaluate GeniWorld from three perspectives. First, GeniWorld achieves superior generative performance over the baselines and demonstrates strong zero-shot interaction modeling across OOD scenarios. Second, across multiple real-world manipulation tasks, policy success rates measured in GeniWorld correlate positively with real-world performance, supporting its utility as a scalable policy evaluator. Third, given a limited set of real demonstrations, GeniWorld generates rich and varied manipulation data that improve downstream policy performance under novel layouts and in complex visual environments.

Our main contributions are summarized as follows:
\begin{itemize}
    \item We introduce GeniWorld, an interactive world model that learns from fixed-scene demonstrations and generalizes robustly to diverse unseen scenarios, supporting robot manipulation in rich synthesized imagination spaces.

    \item We propose an autoregressive generative model conditioned on visual actions to decouple embodiment motion from scene dynamics, enabling explicit interaction modeling and closed-loop interaction with both robot policies and human teleoperators.

    \item We demonstrate that GeniWorld serves as a robust policy evaluator and enables the synthesis of rich, varied manipulation data from limited real-world demonstrations, thereby enhancing downstream policy performance across diverse robotic systems.
\end{itemize}

\begin{figure*}[!t]
    \centering
    \includegraphics[width=\textwidth]{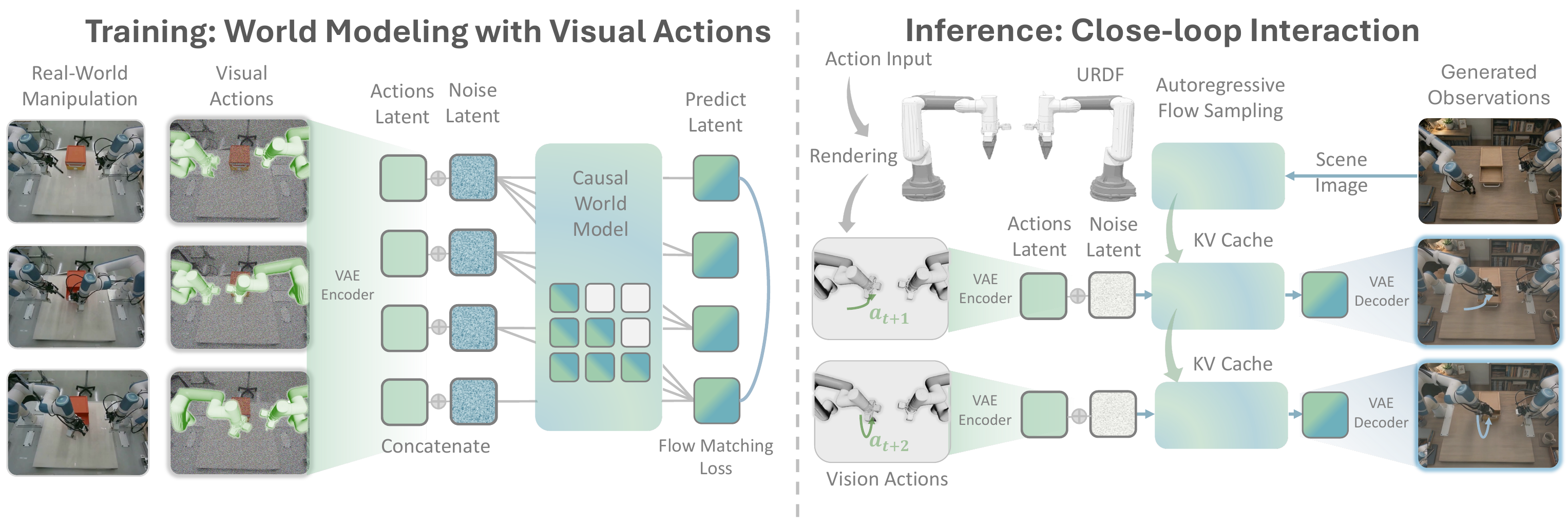}
    \vspace{-1.5em}
    \caption{\textbf{Overview of the GeniWorld method.} We convert robot actions into visual motions via URDF rendering. These motions are subsequently encoded into latent representations and concatenated with noisy video latents. The combined representation serves as input to a causal DiT that predicts future videos via flow matching. During inference, the initial scene image is provided as the first frame. Actions are passed through the URDF-based renderer to produce motion conditions, and the model predicts future observations while leveraging KV caching to maintain historical context.}
    \vspace{-0.3em}
    \label{fig:method}
\end{figure*}

%% file: sections/2-related.tex
\section{Related Work}
\label{sec:related_work}

\paragraph{Action-Conditioned World Models for Robot Manipulation}
Action-conditioned world models, which predict future observations given control actions, have emerged as a pivotal paradigm in robot learning~\cite{goodmanWorldModelsDavid, hafnerDreamControlLearning2020, houWorldModelRobot2026}. Recent advances further scale this paradigm by leveraging video models to simulate high-fidelity visual interactions and environmental changes~\cite{bruce2024genie, assran2025v, nvidiaCosmosWorldFoundation2025, 251215840LargeVideo, zhuUnifiedWorldModels2025a, wu2024ivideogpt, team2025evaluating, yang2023learning}. Existing efforts for robotics~\cite{zhu2025irasim, guo2025ctrlworld, quevedo2025worldgym, li2025worldeval} adapt these video generative models into physical simulators by incorporating robot control actions as conditioning signals for manipulation. However, such methods commonly represent robot motion using implicit, low-dimensional numerical vectors. Passing these vectors directly to the video backbone conflates robot kinematics with scene dynamics, ultimately hindering the model from learning precise, interaction-dependent environmental transitions~\cite{gao2026sword}.

\paragraph{Pixel-Grounded Visual Action Representations}
Many works inject action signals into video generative models by introducing numerical joint or end-effector parameters through AdaLN or cross-attention layers~\cite{yang2023learning, zhu2025irasim, guo2025ctrlworld, quevedo2025worldgym, li2025worldeval, jiang2026wovr, zhu2025wmpo}; however, these implicit representations struggle to capture 3D spatial dynamics accurately. To address this, subsequent works incorporate explicit pixel-aligned conditions, such as projected end-effector poses~\cite{jiang2025enerverseac, yang2026ea, qiu2026ge} or skeletal trajectories~\cite{wang2025vap, wu2026oscar}. Another line of work leverages visual abstractions, such as segmentation masks~\cite{chen2026bridgev2w} and 3D point clouds~\cite{xu2026kinema4d} to provide spatial grounding for video synthesis. However, such explicit representations often fail to capture fine-grained geometric details and structure-grounded interactions. 
Concurrent works~\cite{alzayer2026masked, kim2026robot, bai2026ioi} generate manipulation videos conditioned on URDF renderings of complete trajectories. In contrast, GeniWorld uses visual motion as conditioning signals for causal prediction, enabling closed-loop interaction modeling with the environment. More importantly, we emphasize generalization to unseen scenes and demonstrate the utility of the learned world model for downstream policy evaluation and improvement beyond in-domain settings.
% Concurrent works~\cite{alzayer2026masked, kim2026robot, bai2026ioi} also represent robot actions in visual space through URDF renderings. Our work distinguishes itself through a closed-loop world model that generalizes to unseen scenes, with demonstrated utility for downstream policy evaluation and improvement beyond in-domain settings.
% Recent works~\cite{alzayer2026masked, kim2026robot, bai2026ioi} also represent robot actions in visual space through URDF renderings. Our work advances this direction by developing a URDF-conditioned closed-loop world model that generalizes to unseen scenes, and demonstrates its utility for downstream policy evaluation and improvement beyond in-domain settings.
% Recent works~\cite{alzayer2026masked, kim2026robot, bai2026ioi} also represent robot actions in the visual space through URDF renderings. Our work advances it to a closed-loop world model conditioned on URDF-based robot renderings, with an emphasis on generalization to unseen scenes, and demonstrates its utility for downstream policy evaluation and improvement beyond in-domain settings.
% Crucially, physical interactions stem from environmental dynamics induced by the robot's spatial motion, which align naturally with the strong spatiotemporal dynamics learned by pretrained video generative models~\cite{chu2026wan}. Therefore, directly embedding embodiment motion into future-frame prediction allows the model to faithfully reflect interaction-driven environmental changes.

\paragraph{Data Synthesis for Policy Improvement}
Data synthesis is an effective paradigm for enhancing robot policies. One line of work constructs real-to-sim environments to enable data synthesis across diverse scenes without modifying physical setups~\cite{li2024robogsim, zhao2026high, yang2025novel, gu2025igen}. However, these approaches require complex simulation setups and inevitably suffer from the reality gap. Alternatively, other methods employ generative models for visual augmentation of existing action trajectories~\cite{ye2025anchordream, alhaija2025cosmos, yuan2025roboengine}. Nevertheless, these augmented videos do not support controllable interactions with the environment, thereby limiting the generation of novel robot behaviors. World models enable scalable data synthesis within imagination spaces to improve manipulation-policy performance~\cite{guo2025ctrlworld, quevedo2025worldgym, wang2026interactive}. However, existing methods are largely restricted to in-distribution data generation and remain confined to seen environments and familiar objects. Several methods leverage pretraining on large-scale robot datasets to synthesize richer robot-specific data~\cite{gao2026dreamdojo, jang2025dreamgen, teamEvaluatingGeminiRobotics2026, nvidiaCosmosWorldFoundation2025, team2025evaluating}. However, most robotic systems lack access to such large-scale datasets. Even with limited laboratory-scale data, GeniWorld learns a highly generalizable world model capable of synthesizing diverse manipulation data across novel objects, environments, and spatial layouts. This capability enables broad application across robotic systems for generalist robot learning.

%% file: sections/3-method.tex
\section{Method}
\label{sec:method}
\subsection{Problem Formulation}
\label{sec:problem_formulation}

Given a target robotic system, we aim to learn an interactive world model from an offline dataset $\mathcal{D} = \{\tau_i\}_{i=1}^{N}$ consisting of paired observation--action trajectories $\tau_i = \{(o_0, a_0), (o_1, a_1), \dots, (o_T, a_T)\}$. Given the current observation $o_t$ and an incoming action sequence $a_{t+1:t+H} = (a_{t+1}, \dots, a_{t+H})$, our goal is to construct a world model $\mathcal{W}$ that predicts the future visual observation sequence $o_{t+1:t+H}$:
\begin{equation*}
    o_{t+1:t+H} = \mathcal{W}(o_t, a_{t+1:t+H}).
\end{equation*}
By autoregressively modeling action-sequence-conditioned future prediction, $\mathcal{W}$ enables closed-loop interaction with both human operators and downstream policies.

% \subsection{Decoupling Embodied Dynamics and Interactions}
% \label{sec:visual_action_tokens}

\subsection{World Modeling with Visual Actions}
\label{sec:action_interaction_modeling}

\paragraph{From Numerical to Visual Actions}

We assume that the physical dynamics in manipulation videos are predominantly driven by embodiment kinematics. To this end, we construct the dense visual motion sequence $m_{t+1:t+H}$ by mapping the numerical actions $a_{t+1:t+H}$ through the robot's URDF model and forward-kinematics system and rendering the corresponding embodiment motion from the target camera viewpoint. This rendering isolates the articulated robot structure and excludes objects and background appearance.

We then define the world model as
\begin{equation*}
    o_{t+1:t+H} = \mathcal{W}(o_t, m_{t+1:t+H}),
\end{equation*}
where conditioning on visual action sequences enables the system to learn scene interactions driven directly by embodiment motion.

\paragraph{Spatially Aligned Action Conditioning}
We construct our world model based on a video diffusion backbone~\cite{wan2025wan}. As shown in Fig.~\ref{fig:method}, visual actions are processed via a causal 3D VAE encoder into action latents $z_a \in \mathbb{R}^{C \times L \times H' \times W'}$, which are channel-wise concatenated with the original video latents $z_v \in \mathbb{R}^{C \times L \times H' \times W'}$ to form a spatially aligned joint representation $z = [z_v; z_a] \in \mathbb{R}^{2C \times L \times H' \times W'}$.

This design aligns the embodiment dynamics with the pretrained visual latent space, ensuring strict spatial correspondence between each position in the action and scene latents. By leveraging this action-embedding mechanism, we introduce minimal architectural modifications to the pretrained video backbone, thereby preserving its rich generative priors for interaction modeling~\cite{chu2026wan}.

\paragraph{Causal Modeling}
We formulate future observation prediction as an autoregressive process. We enforce a causal attention mask that restricts each token to attending only to preceding tokens, ensuring that each predicted observation depends strictly on historical states and past action tokens. During training, we initially condition the model on ground-truth frames and subsequently feed model-generated predictions into the observation context to mitigate exposure bias in future prediction~\cite{huang2026self}. 

\paragraph{Training Objective}
Using flow matching~\cite{lipman2022flow}, we predict the next observation conditioned on the history and the corresponding action token. For the next observation latent $z_{t+1}$, the training objective is formulated as follows:

\begin{equation}
\mathcal{L} = \mathbb{E}_{t, s, \mathbf{z}_{t+1}, \boldsymbol{\epsilon}} \left\| \mathbf{v}_\theta \left( \mathbf{z}_{t+1}^{(s)}, s, \mathbf{z}_{\le t} \mid \mathbf{z}_{a,t+1}, \mathbf{c} \right) - \dot{\mathbf{z}}_{t+1}^{(s)} \right\|_2^2 ,
\end{equation}

where $s \in [0, 1]$ denotes the flow timestep, $\mathbf{z}_{t+1}^{(s)} = (1 - s)\boldsymbol{\epsilon} + s\mathbf{z}_{t+1}$ defines the interpolated state with Gaussian noise $\boldsymbol{\epsilon} \sim \mathcal{N}(\mathbf{0}, \mathbf{I})$, and $\dot{\mathbf{z}}_{t+1}^{(s)} = \mathbf{z}_{t+1} - \boldsymbol{\epsilon}$ specifies the ground-truth velocity field. Here, $\mathbf{z}_{\le t}$ represents the historical observation token sequence, $\mathbf{z}_{a,t+1}$ denotes the visual action latent, and $\mathbf{c}$ denotes the language instruction. Only the observation latent is noised and predicted; the visual action serves as a clean conditioning signal.

\input{tables/table1_world_modeling}

\subsection{Efficient Closed-Loop Interaction}
\label{sec:autoregressive_inference}

\paragraph{Autoregressive Prediction}
To enable closed-loop interaction with human operators and downstream policies, the world model predicts the next observation in response to input actions and autoregressively feeds generated observations back as historical context for subsequent rollouts. Following~\cite{li2026causal}, we cache the key-value pairs of preceding tokens as historical context while applying full self-attention within the newly generated token block at each prediction.

\paragraph{Interaction via Motion Rendering}

For both human teleoperators and robot policies, the executed actions are first rendered as dense visual motions using URDF-based forward kinematics aligned with the camera viewpoint. These motions serve as conditioning signals for the world model. Conditioned on the visual action and historical context, the world model synthesizes the corresponding future observation, which is fed back to the agent to guide subsequent decision-making in a continuous control loop.

%% file: tables/table1_world_modeling.tex
\begin{table*}[t]
    \vspace{1.5em}
    \centering
    \caption{
        Quantitative evaluation of world-model performance on the RoboTwin benchmark across Clean-to-Clean and Clean-to-Random settings.
        $\uparrow$ ($\downarrow$) indicates that higher (lower) is better.
        Best results are shown in bold.
    }
    \label{tab:world_modeling_main}
    \scriptsize
    \setlength{\tabcolsep}{1.6pt}
    \renewcommand{\arraystretch}{1.0}
    \resizebox{0.94\textwidth}{!}{%
    \begin{tabular}{lcccccccccccc}
        \toprule
        \multirow{2}{*}{\shortstack[c]{Method /\\action representation}}
        & \multicolumn{6}{c}{Clean-to-Clean}
        & \multicolumn{6}{c}{Clean-to-Random} \\
        \cmidrule(lr){2-7}\cmidrule(lr){8-13}
        & LPIPS$\downarrow$ & PSNR$\uparrow$ & SSIM$\uparrow$ & FID$\downarrow$ & FVD$\downarrow$ & EWMScore$\uparrow$
        & LPIPS$\downarrow$ & PSNR$\uparrow$ & SSIM$\uparrow$ & FID$\downarrow$ & FVD$\downarrow$ & EWMScore$\uparrow$ \\
        \midrule
        Ctrl-World
        & 0.165 & 18.65 & 0.879 & 9.62 & 15.52 & 58.74
        & 0.285 & 20.41 & 0.791 & 21.66 & 35.85 & 51.47 \\
        IRASim
        & 0.323 & 13.26 & 0.801 & 42.98 & 48.98 & 48.83
        & 0.766 & 8.15 & 0.476 & 174.52 & 191.26 & 46.41 \\
        EnerVerse-AC
        & 0.378 & 10.89 & 0.780 & 40.35 & 54.28 & 52.92
        & 0.751 & 12.25 & 0.433 & 74.55 & 99.14 & 31.28 \\
        \midrule
        Ours w/ numerical actions
        & 0.377 & 11.59 & 0.784 & 42.69 & 48.03 & 57.69
        & 0.366 & 13.74 & 0.696 & 40.91 & 53.69 & 51.49 \\
        Ours w/ EE trajectory
        & 0.185 & 17.17 & 0.863 & 12.19 & 18.35 & 60.00
        & 0.300 & 14.17 & 0.716 & 37.69 & 45.89 & 61.36 \\
        Ours w/ skeleton
        & 0.180 & 17.38 & 0.861 & 12.49 & 18.82 & 60.04
        & 0.257 & 18.48 & 0.796 & 31.78 & 39.88 & 61.61 \\
        Ours w/ ControlNet-style conditioning
        & 0.094 & 23.99 & 0.927 & 8.82 & 12.36 & 59.26
        & 0.416 & 14.51 & 0.625 & 40.02 & 59.95 & 56.13 \\
        \midrule
        \textbf{Ours (visual actions)}
        & \textbf{0.055} & \textbf{27.57} & \textbf{0.942} & \textbf{5.59} & \textbf{7.59} & \textbf{61.80}
        & \textbf{0.144} & \textbf{22.71} & \textbf{0.873} & \textbf{13.08} & \textbf{20.15} & \textbf{63.54} \\
        \bottomrule
    \end{tabular}
    }
\end{table*}

\begin{figure*}[t]
    \centering
    \includegraphics[width=\textwidth]{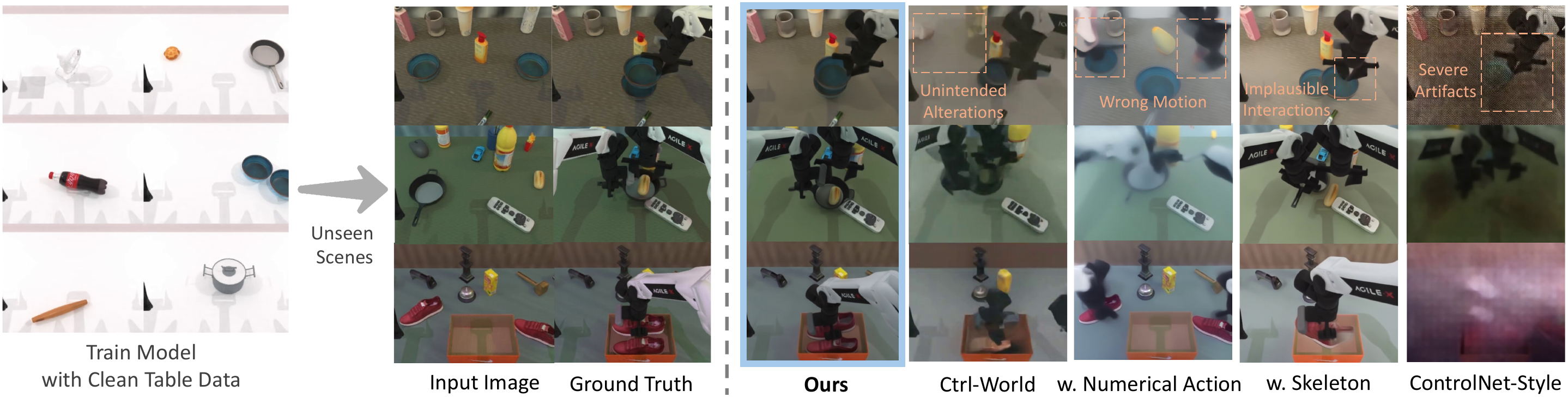}
    \caption{\textbf{Qualitative Clean-to-Random world-modeling results.}
    All models are trained on clean tabletop data and evaluated in unseen scenes. GeniWorld preserves the commanded robot motion and produces interaction outcomes that closely match the ground truth, whereas competing methods fail to generate the expected results in out-of-distribution (OOD) scenarios. Furthermore, the ablation results show that our dense visual actions enable more accurate interaction modeling.}
    \vspace{-1.5em}
    \label{fig:world_modeling_qualitative}
\end{figure*}

% \begin{figure*}[t]
%     \centering
%     \begin{minipage}[t]{\columnwidth}
%         \vspace{0pt}
%         \centering
%         \includegraphics[width=\linewidth]{figures/PPT_Paper_Figures_4.pdf}
%         \captionof{figure}{Training convergence. Visual action conditioning
%         converges faster than numerical action conditioning.}
%         \label{fig:efficiency}
%     \end{minipage}
%     \hfill
%     \begin{minipage}[t]{\columnwidth}
%         \vspace{0pt}
%         \centering
%         \includegraphics[width=\linewidth]{figures/PPT_Paper_Figures_5.pdf}
%         \captionof{figure}{Few-step inference. Visual action conditioning
%         maintains higher world-modeling quality with fewer denoising steps.}
%         \label{fig:few-step-inference}
%     \end{minipage}
% \end{figure*}

%% file: sections/4-experiment.tex
\section{Experiments}
\label{sec:experiments}
In this section, we design comprehensive experiments to evaluate GeniWorld by addressing the following questions:
\begin{enumerate}
    \item \textbf{Generative Fidelity \& Generalization:} Can GeniWorld generate high-fidelity video predictions and generalize effectively across diverse, unseen scenarios?
    \item \textbf{Policy Evaluation Reliability:} Can GeniWorld reliably evaluate downstream policies while maintaining a strong correlation with real-world performance under scene perturbations and out-of-domain settings?
    \item \textbf{Data Synthesis for Policy Improvement:} Can GeniWorld synthesize effective training data from limited recorded trajectories to improve policy learning and performance without the cost of constructing new real-world setups?
\end{enumerate}

%%%%%%%%%%%%%%%%%%%%%%%%%%%%%%%%%%%%%%%%%%%%%%%%%%%%%%%%%%%%%%%%%%%%%%%%%%%%%%%%
\subsection{World Modeling Quality}
\label{sec:world_modeling_quality}

\paragraph{Experimental Setup}

Following~\cite{shang2026worldarena}, we establish two settings for evaluating
action-conditioned world modeling based on RoboTwin2.0~\cite{chen2025robotwin}. The \textbf{Clean-to-Clean} setting trains and evaluates on the Clean dataset. The Clean dataset are split, with 45 trajectories per task used for training and 5 used for evaluation. Across 50 tasks, this produces 2,250 training episodes and 250 held-out
test episodes.
To evaluate out-of-distribution generalization, we introduce
the \textbf{Clean-to-Random} setting for zero-shot out-of-distribution evaluation. Models are
trained on the same 2,250 Clean episodes and evaluated, without
adaptation, on 250 episodes from the Random subset. These episodes
introduce substantial variations in scene appearance, object instances,
object placements, and task-relevant spatial layouts. Each episode
contains 121 frames rendered at 24~fps. In both settings, the model receives an initial observation $I_0$, a
language instruction $\ell$, and a future action trajectory
$a_{0:T-1}$, and predicts the corresponding future observations.

We evaluate visual prediction quality using PSNR, SSIM~\cite{wang2004image}, LPIPS~\cite{zhang2018unreasonable}, FID~\cite{heusel2017gans}, FVD~\cite{unterthiner2018towards} and EWMScore~\cite{shang2026worldarena}. We compare GeniWorld
with Ctrl-World~\cite{guo2025ctrlworld}, IRASim~\cite{zhu2025irasim}, and EnerVerse-AC~\cite{jiang2025enerverseac} under
the same training split. For the ablation analysis, we evaluate four action-conditioning designs built on the same video backbone: numerical actions, projected end-effector poses, projected skeleton actions, and our dense visual actions. We further compare our concatenation-based action embedding with ControlNet-style conditioning~\cite{zhang2023adding}.

% The controlled comparison under the shared Wan2.2-TI2V-5B backbone
% further isolates the effect of the action representation. Replacing
% numerical actions with projected skeletons improves the Random score
% from 44.18 to 48.31, while replacing skeletons with dense visual masks
% further increases the score to 52.67. These results indicate that the advantage of visual
% action conditioning becomes more pronounced under distribution shift.
% By representing robot motion directly in the observation space, visual
% masks reduce the burden of learning the mapping from numerical control
% signals to view-dependent image dynamics and provide a more
% transferable interface for modeling robot--environment interactions.

%%%%%%%%%%%%%%%%%%%%%%%%%%%%%%%%%%%%%%%%%%%%%%%%%%%%%%%%%%%%%%%%%%%%%%%%%%%%%%%%

\vspace{0pt}
\paragraph{World-Model Quality and OOD Generalization}
As shown in Table~\ref{tab:world_modeling_main}, GeniWorld achieves the
highest generation quality in both the Clean-to-Clean and
Clean-to-Random settings. In the in-domain evaluation, our method consistently achieves superior performance across all six metrics. Compared to ablation baselines using numerical action control or alternative explicit representations (e.g., end-effector pose and skeleton), our model demonstrates higher alignment with ground-truth observations. Furthermore, under domain shifts from clean to randomized environments, prior works such as IRASim experience severe degradation, with FID and FVD dropping to $174.52$ and $191.26$, respectively. In contrast, GeniWorld maintains remarkable fidelity, retaining low FID and FVD scores of $13.08$ and $20.15$. These results demonstrate that GeniWorld achieves superior generation quality even when scene appearance, object instances, object placements, and task-relevant spatial layouts differ significantly from those seen during training.

As illustrated in Fig.~\ref{fig:world_modeling_qualitative}, despite being trained solely on a fixed, plain scene, GeniWorld synthesizes accurate interactions and dynamics across novel randomized scenes, successfully executing target tasks guided by input actions. In contrast, baseline methods such as Ctrl-World suffer from severe visual artifacts in complex textures and cause unintended changes in task-irrelevant objects during manipulation. Furthermore, implicit numerical actions fail to generate accurate robot motions in unseen environments, while skeleton-based control often generates physically implausible interactions, such as contactless grasping.

\begin{figure}[h]
    \centering
    \includegraphics[width=0.99\linewidth]
    {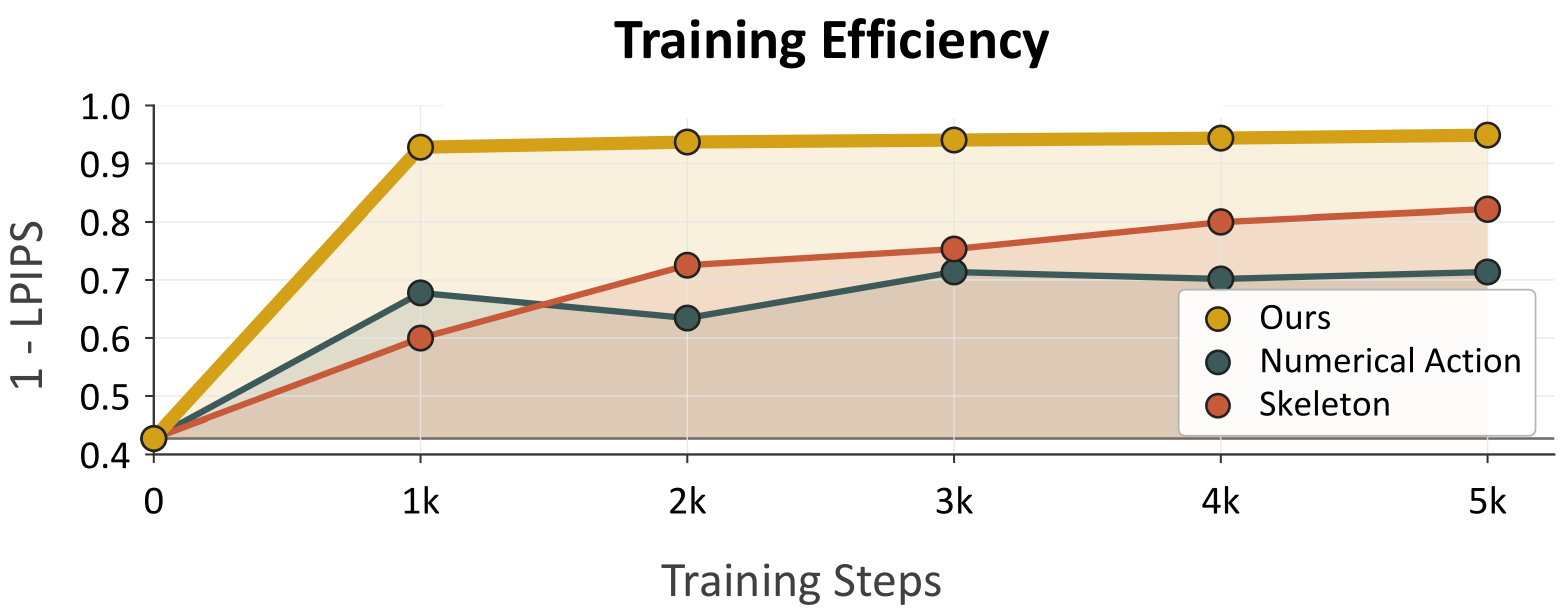}
    \vspace{-2.0em}
    \caption{\textbf{Training convergence across different action representations.} Under identical training configurations, we compare generation quality ($1 - \text{LPIPS}$; $y$-axis) against the number of training steps ($x$-axis) for various action representations.}
    \vspace{-0.5em}

    \label{fig:efficiency}
\end{figure}

\paragraph{Training Efficiency}
We evaluate the training efficiency of our visual action representation
under identical backbone architectures and experimental setups. Fig.~\ref{fig:efficiency}
presents the evaluation metrics across training iterations for different
action input modalities. Compared to baseline representations, visual actions achieve faster convergence and consistently superior performance, producing high-fidelity outputs even in early training stages and consistently outperforming other action representations. This demonstrates that visual actions provide a more effective
input modality by leveraging the generative priors of pretrained video models.

\begin{figure}[h]
    \centering
    \vspace{0.3em}
    \includegraphics[width=0.97\linewidth]{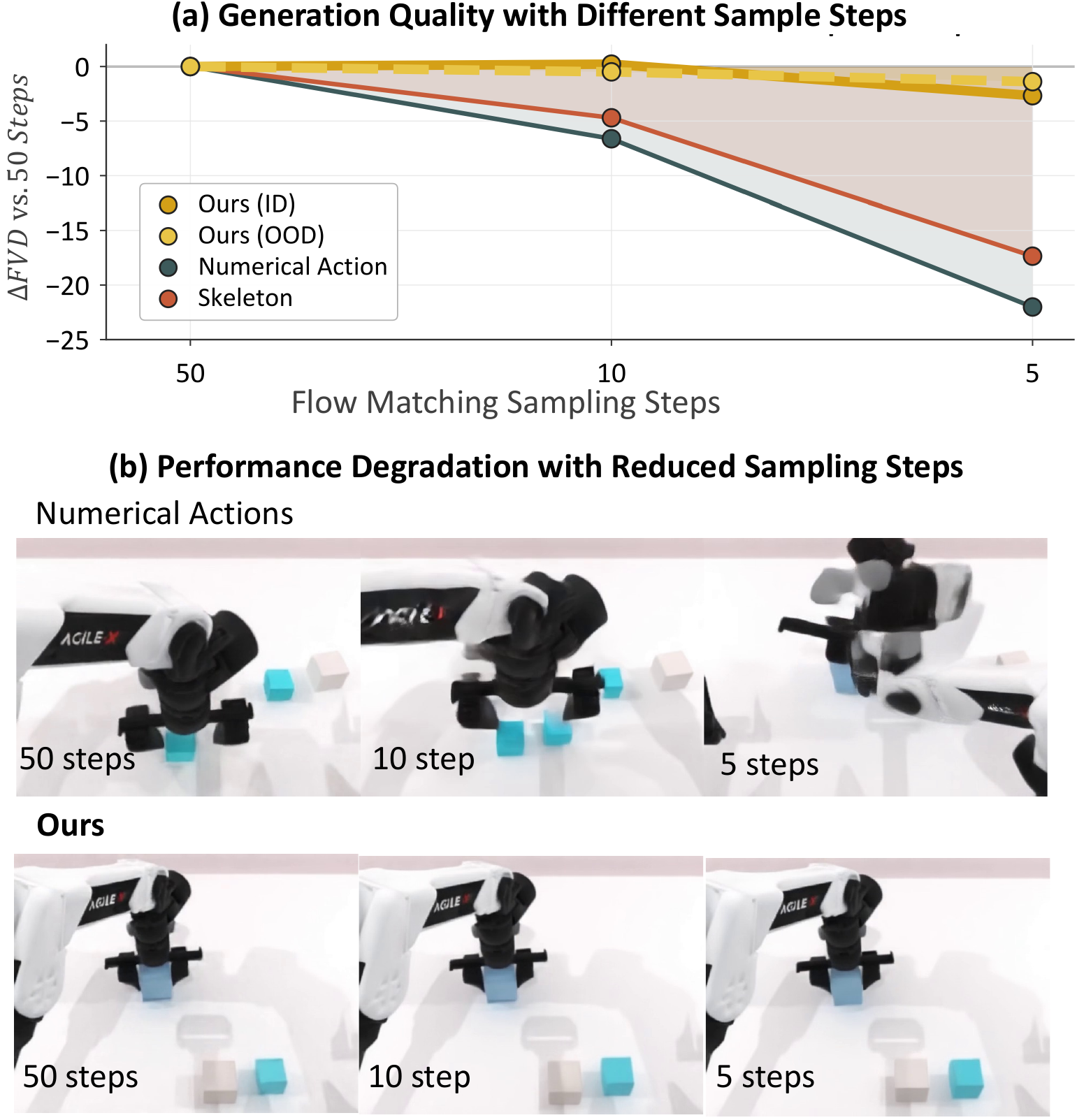}
    \vspace{0.3em}
    \caption{\textbf{Performance with different numbers of flow-matching sampling steps.} We evaluate world-modeling quality using 50, 10, and 5 inference steps, where fewer steps correspond to faster inference. Visual action conditioning consistently maintains superior generation quality with fewer denoising steps than the other action representations.}
    \label{fig:few-step-inference}
    \vspace{-0.5em}
\end{figure}

\paragraph{Inference Efficiency}
For robotic manipulation tasks, we observe that visual action conditioning remains highly effective with substantially fewer flow-matching sampling steps. As shown in Fig.~\ref{fig:few-step-inference}, reducing the number of sampling steps from 50 to 10 or even 5 incurs negligible quality degradation across both in-domain and out-of-domain scenarios. Specifically, our model achieves a $\sim 10\times$ inference speedup with an FVD degradation of only $\sim 2\%$, whereas the numerical-action-conditioned baseline deteriorates considerably with an FVD drop of $\sim 22\%$. Furthermore, as illustrated in Fig.~\ref{fig:few-step-inference}(b), visual action conditioning provides strong motion priors for interactive dynamics generation; in contrast, numerical action conditioning suffers severe visual degradation under few-step sampling. This demonstrates that our design effectively increases the interaction rate.

\subsection{Real-World Platform Setup}

\begin{figure}[h]
    \centering
    \includegraphics[width=0.99\linewidth]{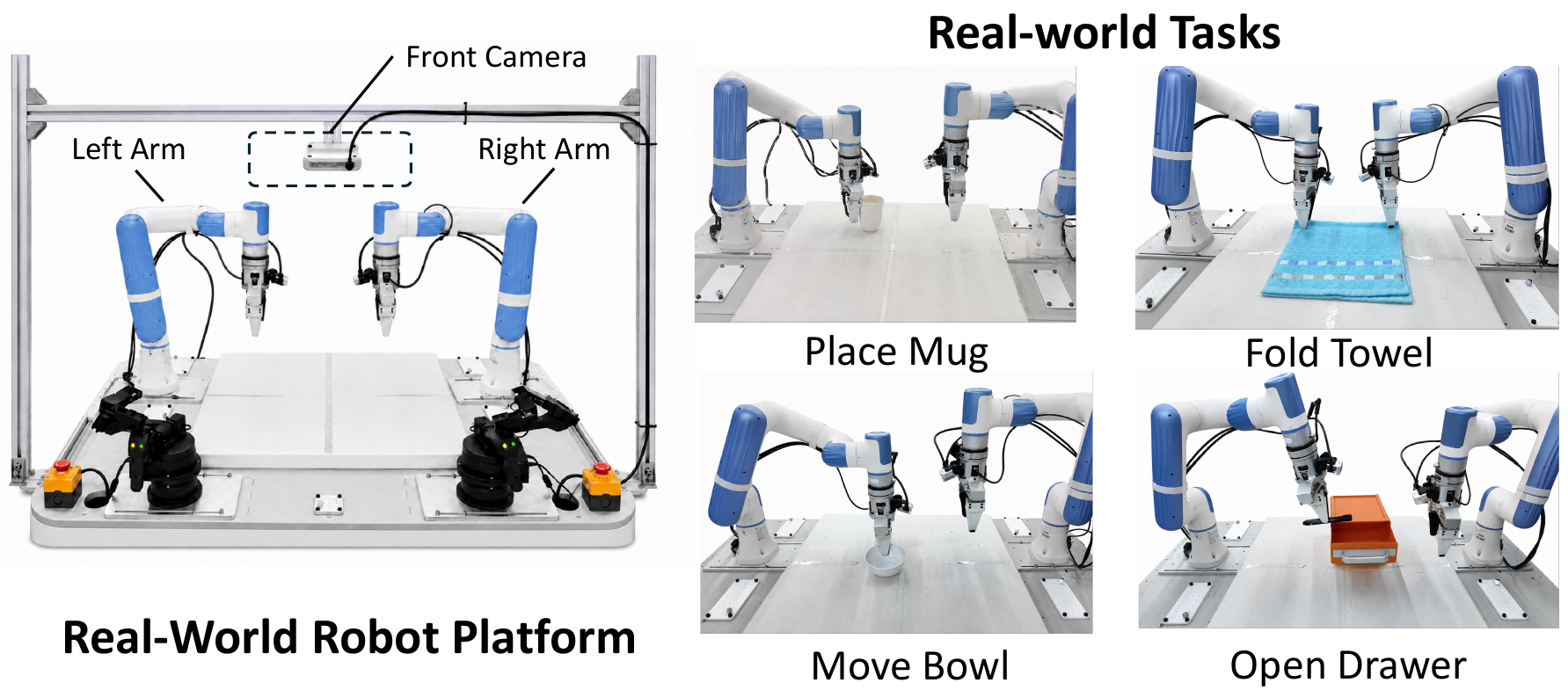}
    \vspace{-0.3em}
    \caption{\textbf{Real-world experimental setup.} Our platform uses a dual-arm Xtrainer robotic system and includes four manipulation tasks.}
    \label{fig:real_world_setup}
\end{figure}

We construct our experimental platform using a dual-arm Xtrainer robotic system as shown in Figs.~\ref{fig:real_world_setup}. Following hand--eye calibration, we integrate the robot's URDF into Isaac Sim for motion control and replicate the physical camera setup to render visually consistent observations.

We design four representative manipulation tasks involving diverse objects and complex physical dynamics: \textbf{Move Bowl}, \textbf{Fold Towel}, \textbf{Place Mug}, and \textbf{Open Drawer}. Demonstrations are collected using the Xtrainer teleoperation system. During training-data collection, we maintain a clean tabletop environment and randomize only the poses of the manipulated objects across episodes.

\subsection{World Model for Policy Evaluation}
\label{sec:policy_evaluation}
We investigate whether GeniWorld can serve as an effective offline evaluator for downstream policies. By executing the same policy checkpoints in both GeniWorld and real-world setups, we quantitatively measure the correlation between simulated rollouts and physical task success rates.

\begin{figure}[h]
    \centering
    \includegraphics[width=0.95\linewidth]{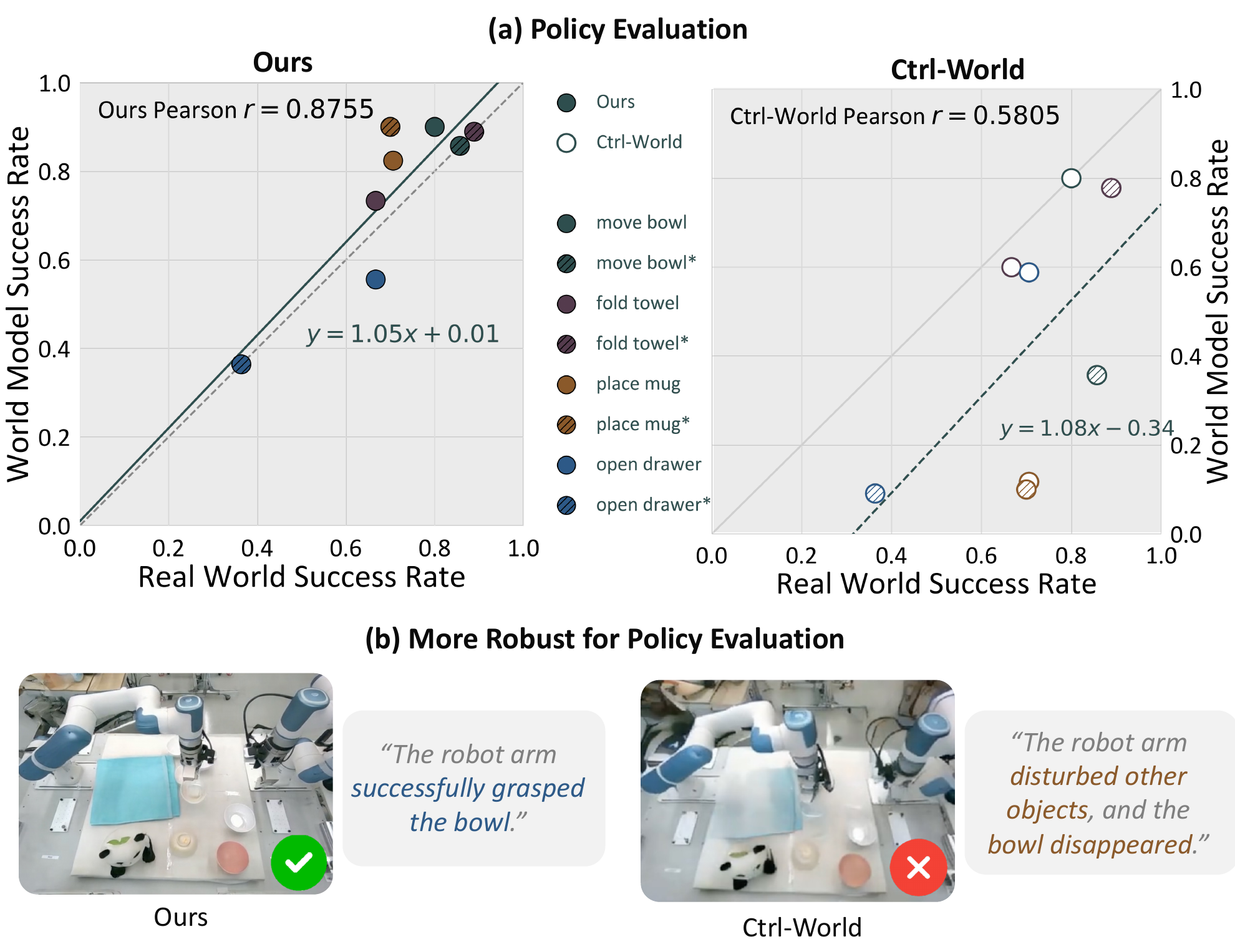}
        \vspace{-0.0em}
        \captionof{figure}{\textbf{Correlation between policy performance in GeniWorld and the real world.} (a) Joint evaluation of policy performance in in-domain scenarios and OOD scenes with visual distractions. Here, * denotes tasks with environmental distractions. (b) Comparison showing that our world simulator is more robust in OOD scenarios.}
        \label{fig:policy_correlation}
\end{figure}

Policies are fine-tuned from the $\pi_0$ vision-language-action (VLA) model~\cite{black2024pi_0} using the collected real-world demonstrations. For each trial, the policy is initialized with the same first-frame visual observation in GeniWorld and the real-world environment. Human evaluators assess binary task success, while a vision-language model (VLM)~\cite{bai2025qwen3, achiam2023gpt} serves as an automated judge of manipulation plausibility.

Fig.~\ref{fig:policy_correlation}(a) plots real-world success rates ($x$-axis) against success rates predicted by the world models ($y$-axis) across the evaluated tasks. \textbf{GeniWorld} exhibits a strong positive correlation between real-world and simulated performance in standard in-domain settings. This alignment remains consistent under environmental perturbations, demonstrating the model's stability in noisy conditions. As illustrated in Fig.~\ref{fig:policy_correlation}(b), environmental perturbations severely corrupt the generated interactions involving the manipulated object in Ctrl-World. In contrast, our model is substantially more robust to visual distractors and consistently generates accurate manipulation observations. These results demonstrate that our world model provides reliable policy evaluation under diverse conditions.

\subsection{Data Synthesis for Policy Improvement}
\label{sec:policy_improvement}

We next ask whether \textbf{GeniWorld} can leverage limited demonstrations from a narrow distribution to synthesize diverse manipulation data, thereby improving policy performance and generalization under complex conditions.

\paragraph{Experimental Setup}

We train the $\pi_0$ VLA model on the four tasks \textbf{using only 25 demonstrations} per task. In addition to the standard in-domain setting, we evaluate task execution under challenging out-of-distribution (OOD) conditions: (i) \textbf{Spatial rearrangement}, where target-object positions are randomized over a broader area; (ii) \textbf{Novel object instances}, where target objects are replaced with unseen instances; (iii) \textbf{Visual distractors}, where random objects are placed on the tabletop; and (iv) \textbf{Lighting shifts}, where ambient illumination is varied.

\begin{figure}[h]
    \centering
    \includegraphics[width=0.90\linewidth]{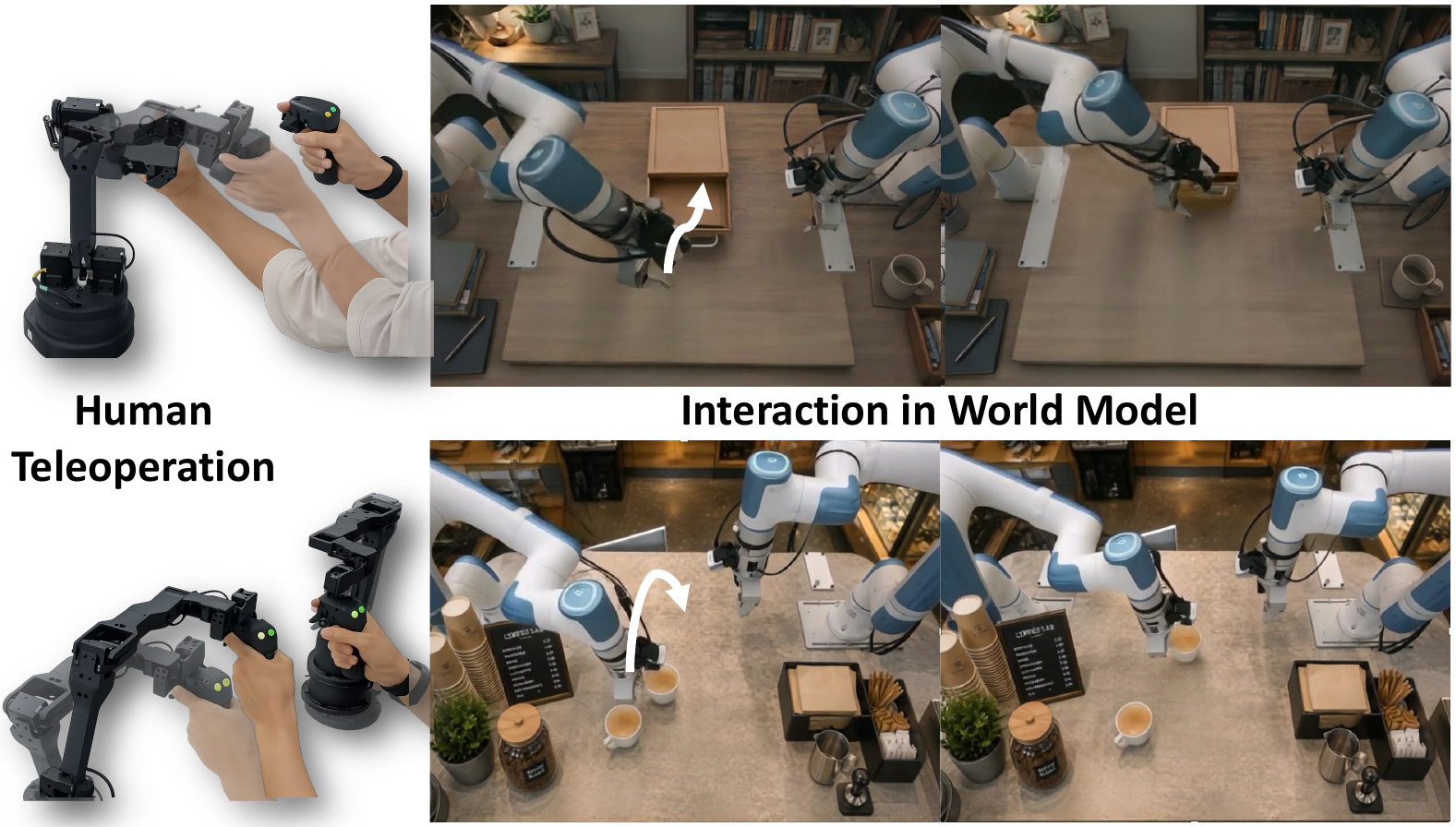}
    \vspace{0.0em}
    \caption{\textbf{Teleoperation with the interactive world model.} A human operator teleoperates the robot in real time based on the generated scene, while the model autoregressively generates the corresponding visual responses.}
    \label{fig:teleoperation}
\end{figure}

\begin{figure*}[!t]
    \centering
    \includegraphics[width=\textwidth]{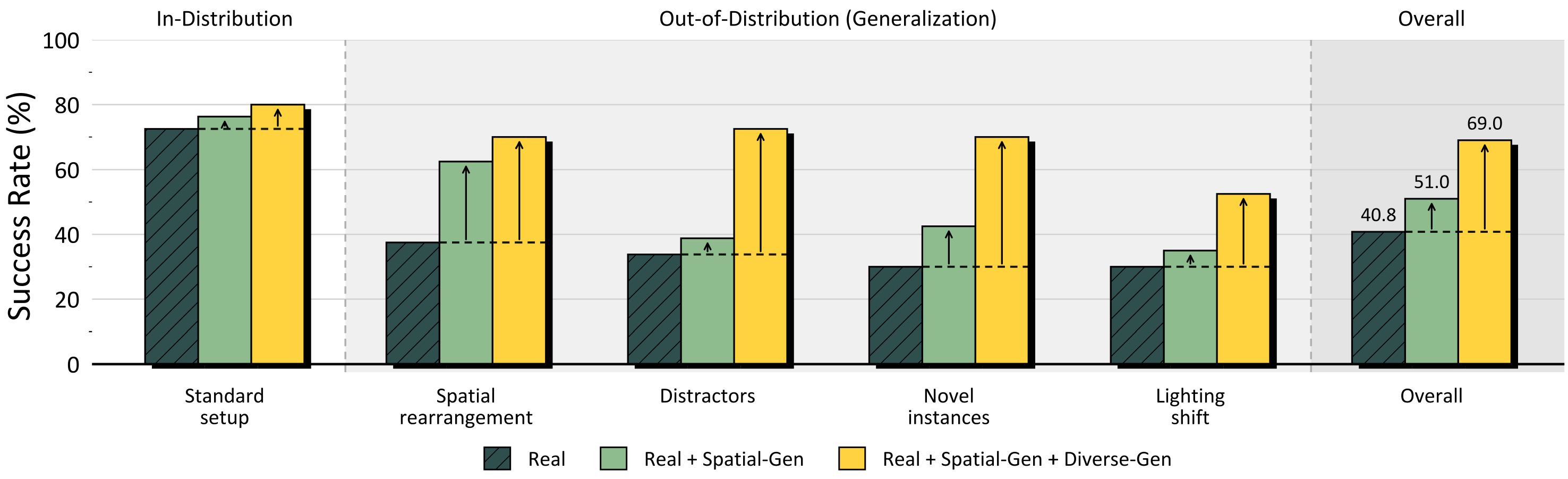}
    \vspace{-0.5em}
    \caption{\textbf{Real-world policy success rates across different settings.} \textbf{Real} denotes policies trained strictly on real-world data; \textbf{Real + Spatial-Gen} adds GeniWorld-synthesized data focused on spatial generalization; and \textbf{Real + Spatial-Gen + Diverse-Gen} adds both spatially randomized data and data with diverse scene variations. The combined regime achieves the strongest overall performance.}
    \label{fig:policy_improvement}
    \vspace{-0.7em}
\end{figure*}

\begin{figure}[h]
    \centering
    \includegraphics[width=0.99\linewidth]{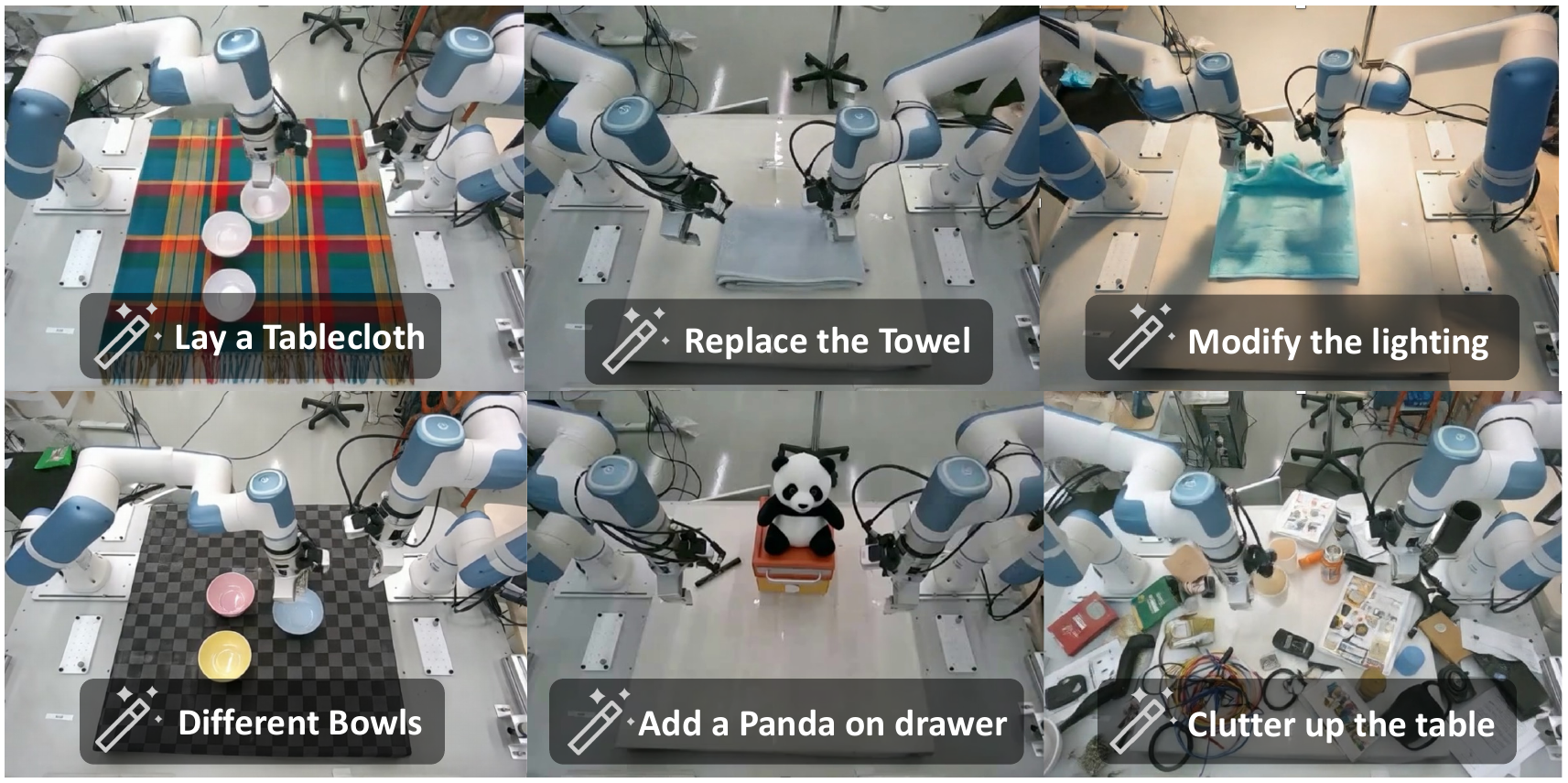}
    \vspace{-0.3em}
    \caption{\textbf{Data generation for real-world robot manipulation.} We use image-editing models to modify the original scene images and create diverse, complex conditions. New manipulation trajectories are synthesized through action replay or teleoperation.}
    \label{fig:gendata}
\end{figure}

\paragraph{Diverse Data Synthesis via World Model}
Using advanced image-generation models (e.g., GPT-Image~\cite{openai2025gptimage} and Qwen-Image~\cite{wu2025qwen}), we synthesize high-variance manipulation scenarios through instruction-driven editing, as shown in Fig.~\ref{fig:gendata}. Specifically, we modify the initial frame using targeted instructions, such as placing target objects in edge-case spatial configurations, introducing complex environmental changes, and replacing target objects. As shown in Figs.~\ref{fig:teleoperation}, we construct a system for teleoperating the world model by linking teleoperation hardware to URDF-based robot kinematics in simulation. The system streams rendered visual-action frames directly to the world model during interactive control. With 5 sampling steps, the model achieves an interactive inference rate of approximately 8~Hz on an NVIDIA H20 GPU. The edited images serve as initial frames for conditioning the world model. For existing trajectories, we replay the recorded action sequences through the simulator. To generate novel behaviors, operators interactively control the world model and synthesize new manipulation trajectories. For each task, we synthesize 65 in-domain trajectories with spatial randomization and 65 diverse trajectories under varied conditions.

\paragraph{Policy Performance Enhancement}

We train the policy under four data regimes for each task: (i) \textbf{Real-Only}, using 25 real-world demonstrations; (ii) \textbf{+ Spatial-Gen}, adding 65 spatially randomized trajectories generated by GeniWorld; and (iii) \textbf{+ Spatial-Gen + Diverse-Gen}, further adding 65 synthetic trajectories generated under diverse environmental conditions. All policy variants are evaluated under identical physical and OOD test settings to ensure a fair comparison.

\begin{figure}[h]
    \centering
    \includegraphics[width=0.95\linewidth]{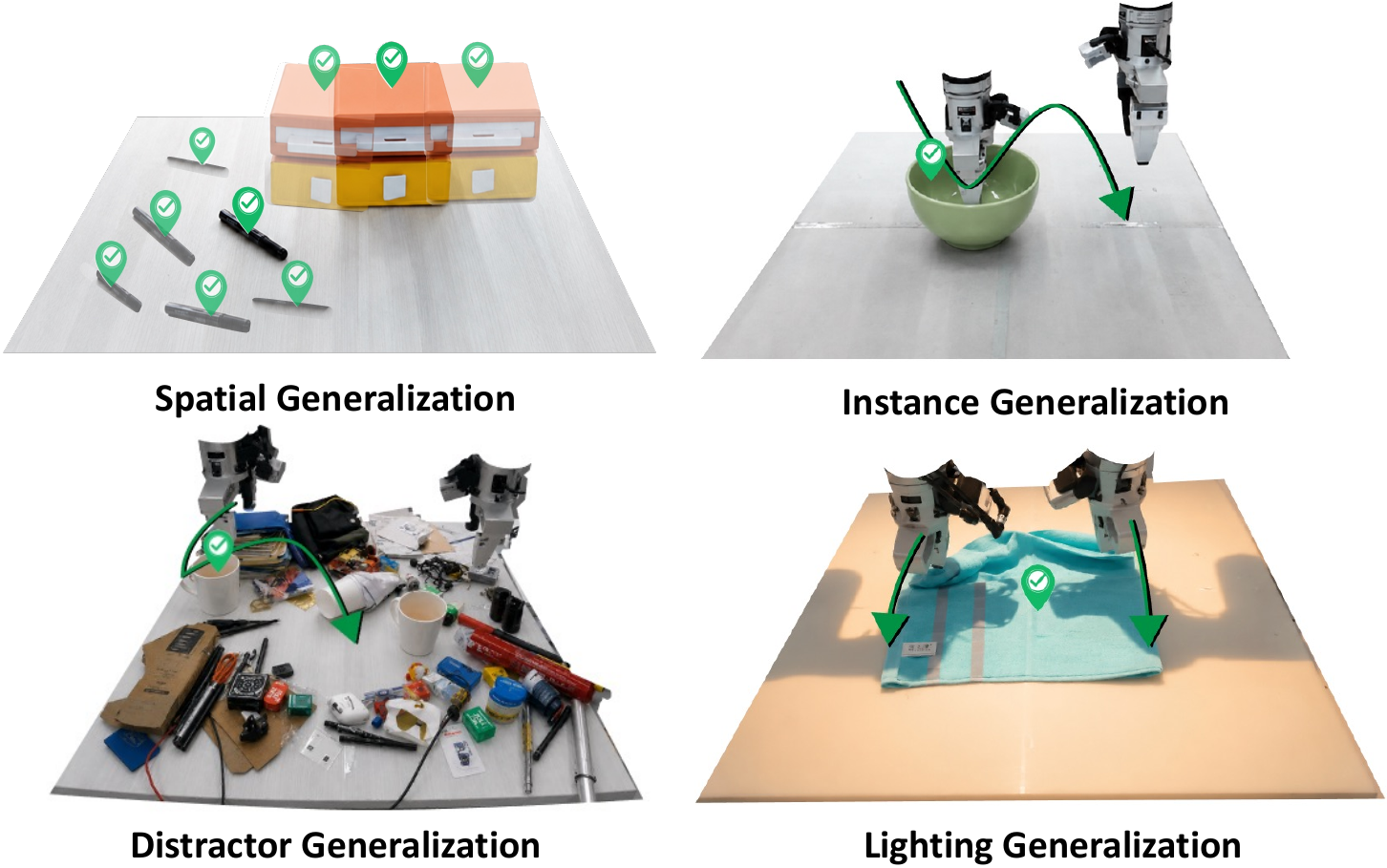}
    \vspace{-0.0em}
    \caption{\textbf{Performance of policies augmented with GeniWorld-synthesized data.} Even with minimal collected real-world data, policies augmented with GeniWorld-synthesized data demonstrate superior performance across diverse OOD conditions.}
    \label{fig:policydemo}
    \vspace{-0.0em}
\end{figure}

Fig.~\ref{fig:policy_improvement} compares the three training regimes. The results shows
that Spatial-Gen and Diverse-Gen provide complementary benefits. Spatial-Gen improves success
under spatial rearrangement from 37.5\% to 62.5\%, compared with 42.5\%
for Diverse-Gen, whereas Diverse-Gen produces larger gains
under distractors (63.8\% vs.\ 38.8\%), novel instances (62.5\% vs.\
42.5\%), and lighting shifts (42.5\% vs.\ 35.0\%). Combining both
sources performs best across all settings, reaching 70.0\% under spatial
rearrangement, 72.5\% with distractors, 70.0\% on novel instances, and
52.5\% under lighting shifts, while increasing the overall success rate
from 40.8\% to 69.0\%. Representative policy behaviors under different OOD conditions are shown in Fig.~\ref{fig:policydemo}.
These results demonstrate that GeniWorld effectively mitigates data scarcity in policy learning by synthesizing highly diverse data without the high cost of manual scene construction and data collection. Consequently, our approach substantially improves overall policy performance in challenging out-of-distribution scenarios.

%% file: sections/5-conclusion.tex
\section{Conclusion}
\label{sec:conclusion}

In this work, we introduce GeniWorld, a generalizable interactive world model conditioned on visual actions that enables spatially grounded and closed-loop robot interaction. Despite training on limited fixed-scene demonstrations, GeniWorld achieves strong in-domain performance and reliable zero-shot generalization to unseen scenarios. It enables robust policy evaluation and generates scalable, highly diverse synthetic trajectories that improve downstream policy performance. These results highlight the strong potential of generalizable world models to provide scalable imagination spaces for embodied robot learning. 

%% file: sections/X_appendix.tex
\clearpage
\appendices

\section{Training Details}
\label{sec:appendix_implementation}

\subsection{Model Architecture \& Training Setup}
Our world model backbone is built upon the Wan2.2-TI2V-5B model and optimized using Flow Matching with an autoregressive causal attention mechanism. To integrate action conditioning, visual actions are encoded into latent representations using the same VAE encoder, matching the visual latent dimensions ($C = 48$). We perform channel-wise concatenation of the action and video latents to form a $96$-channel input ($48+48$). Accordingly, the original 3D convolutional patch embedding layer ($\text{channels}=48$) is replaced with a expanded layer ($\text{channels}=96$). The first $48$ input channels copy the pre-trained weights to preserve the original visual pathways, while the $48$ newly added action channels are initialized using Kaiming initialization and scaled by $0.1\times$ to stabilize early gradient flow. The bias parameters are inherited directly from the original layer.

The diffusion process utilizes $1000$ timesteps with a noise schedule shift parameter of $5.0$. We train the model using a learning rate of $1\times 10^{-5}$ and a global batch size of $4$ on 4 NVIDIA H20 GPUs.. To enhance visual robustness, color overlay augmentation is applied to both the ground-truth video sequences and the initial frame with a probability of $p = 0.5$.

\subsection{Inference \& Autoregressive Generation}
During inference, we employ KV-cache to enable autoregressive prediction. Visual sequences are generated at a spatial resolution of $480 \times 640$, where $121$ raw video frames correspond to $31$ latent frames after VAE encoding. Generation proceeds chunk-by-chunk with a size of $3$ latent frames per autoregressive step, while the initial frame ($I_0$) is processed independently as a single block. We apply Classifier-Free Guidance (CFG) with a scale of $3.0$. A detailed summary of the hyperparameters is provided in Table~\ref{tab:hyperparameters}.

\begin{table}[h]
\centering
\caption{\textbf{Implementation and Hyperparameter Details.}}
\label{tab:hyperparameters}
\begin{tabular}{ll}
\toprule
\textbf{Parameter / Configuration} & \textbf{Value} \\
\midrule
Backbone Architecture & Wan2.2-TI2V-5B \\
Optimization Objective & Flow Matching \\
Learning Rate & $1 \times 10^{-5}$ \\
Global Batch Size & 4 \\
Diffusion Timesteps ($T$) & 1000 \\
Noise Schedule Shift & 5.0 \\
\midrule
Inference Spatial Resolution & $480 \times 640$ \\
Frame Count & 121 pixel frames (31 latent frames) \\
Autoregressive Block Size & 3 latent frames / block \\
Classifier-Free Guidance (CFG) & 3.0 \\
\bottomrule
\end{tabular}
\end{table}

\section{Evaluation Metrics.}

To evaluate visual fidelity and temporal consistency, we report pixel-level quality via frame-wise PSNR and SSIM ($\uparrow$), perceptual distance via a pre-trained AlexNet-backed LPIPS ($\downarrow$), frame-level distribution distance via FID ($\downarrow$) using an ImageNet pre-trained Inception-v3, and video-level temporal distance via FVD ($\downarrow$), which computes Fréchet distance over Inception-v3 features aggregated by concatenated temporal mean and standard deviation.

\section{Details of Policy Improvement}
\label{app:policy_improvement}
\begin{figure*}[!t]
    \centering
    \includegraphics[width=0.8\textwidth]{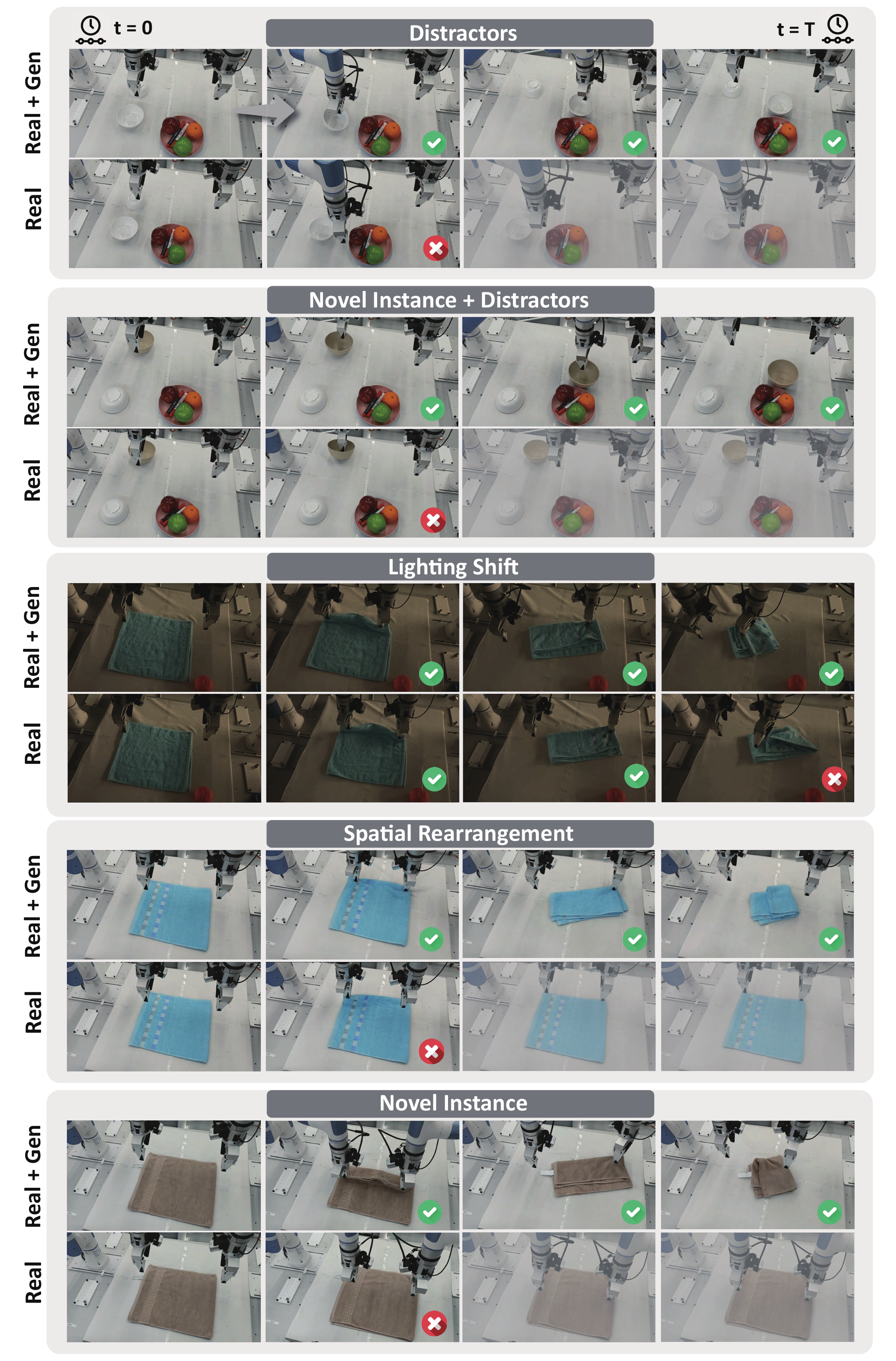}
    % \vspace{-2em}
    \caption{\textbf{Comparison of real-world policy rollouts.} We visualize rollouts of $\pi_0$ policies trained on real data alone and on real data augmented with GeniWorld-generated trajectories. We found that trained on both real and generated data, $\pi_0$ shows a better performance on OOD scenarios such as distractors, novel instances, spatial rearrangement and lighting shift.}
    \vspace{-0.8em}
    \label{fig:rollout1}
\end{figure*}

\begin{figure*}[!t]
    \centering
    \includegraphics[width=0.8\textwidth]{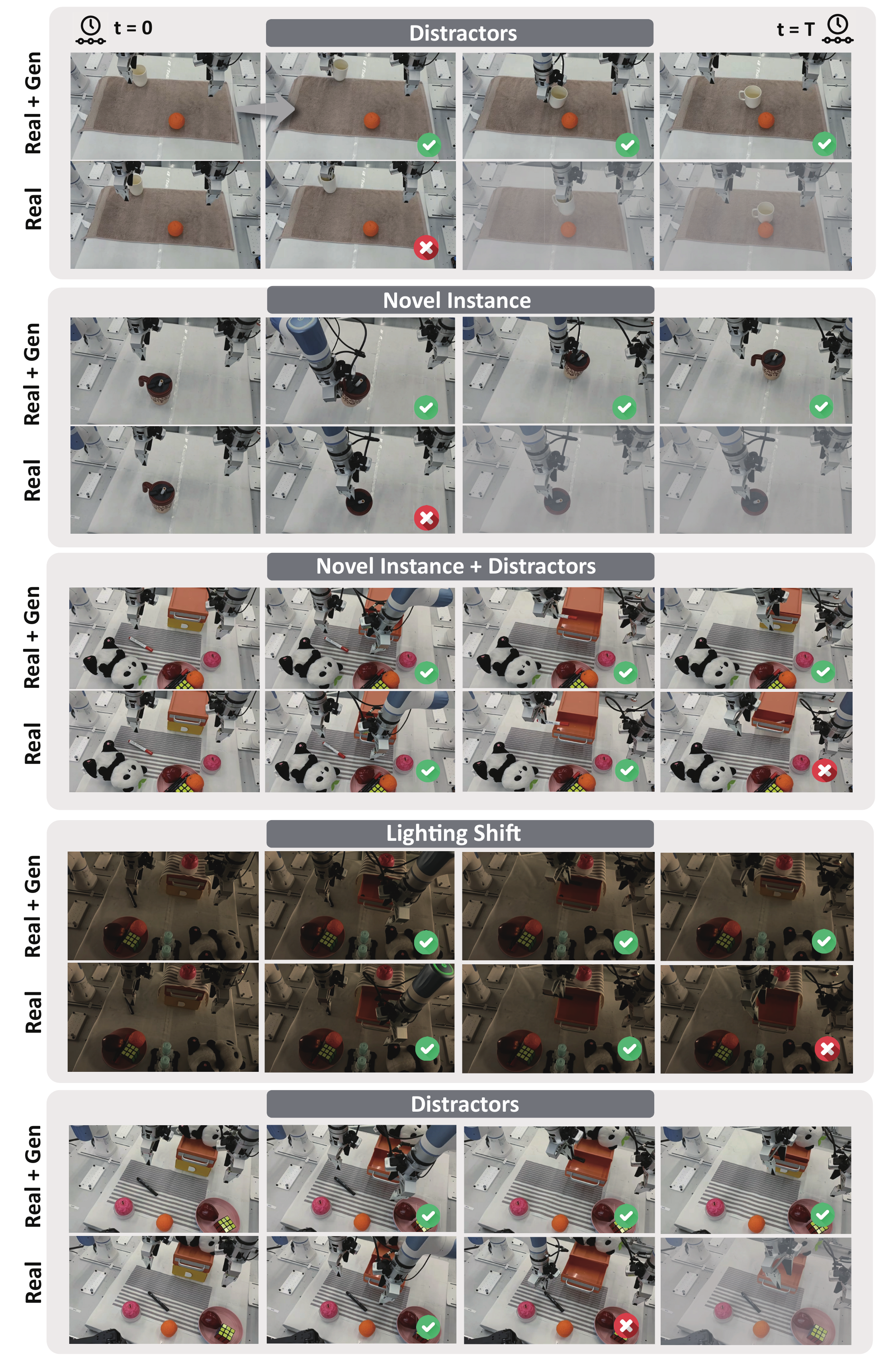}
    % \vspace{-2em}
    \caption{\textbf{More visualization of real-world policy rollouts.} }
    \vspace{-0.8em}
    \label{fig:rollout2}
\end{figure*}

This section provides policy-training details, the evaluation protocol, and per-task results for the policy-improvement experiments. Table~\ref{tab:policy_improvement_per_task} reports real-world success rates for the three training-data regimes across all evaluation settings. Figs.~\ref{fig:rollout1} and~\ref{fig:rollout2} present qualitative comparisons of real-world rollouts from $\pi_0$ policies trained on real data alone and on real data augmented with GeniWorld-generated data.

\paragraph{Policy training details}
For the policy-improvement experiments, we independently fine-tune the
publicly released $\pi_0$ base model using the official OpenPI
implementation. Given an observation $\mathbf{o}_t$ and an action chunk
$\mathbf{A}_t=[\mathbf{a}_t,\ldots,\mathbf{a}_{t+H-1}]$ with $H=16$,
the noisy action input is constructed as
\begin{equation}
    \mathbf{A}_t^\tau
    =
    \tau\mathbf{A}_t+(1-\tau)\boldsymbol{\epsilon},
    \qquad
    \boldsymbol{\epsilon}\sim\mathcal{N}(\mathbf{0},\mathbf{I}),
\end{equation}
where $\tau\in[0,1]$ is the flow timestep. The policy is trained using
the conditional flow-matching objective
\begin{equation}
    \mathcal{L}_{\mathrm{FM}}
    =
    \mathrm{E}
    \left[
    \left\|
    \mathbf{v}_{\theta}(\mathbf{A}_t^\tau,\mathbf{o}_t)
    -
    (\mathbf{A}_t-\boldsymbol{\epsilon})
    \right\|_2^2
    \right].
\end{equation}
All policies are initialized from the same $\pi_0$ checkpoint and use
identical training configurations; only the composition of the training
data differs across regimes.

\begin{table*}[t]
    \centering
    \caption{Per-task real-world policy success rates (\%) with generated training data. Each task is evaluated over 20 trials in each setting. Mean averages the four tasks, while Overall averages the five evaluation settings.}
    \label{tab:policy_improvement_per_task}
    \normalsize
    \setlength{\tabcolsep}{3pt}
    \renewcommand{\arraystretch}{1.10}
    \begin{tabular*}{\textwidth}{@{\extracolsep{\fill}}clcccccc@{}}
        \toprule
        \multirow[c]{2}{*}{\shortstack[c]{Training data}}
        & \multirow[c]{2}{*}{Task ($n=20$)}
        & \multicolumn{5}{c}{Evaluation setting}
        & \multirow[c]{2}{*}{Overall} \\
        \cmidrule(lr){3-7}
        & & Standard & Spatial & Distractors & Novel inst. & Lighting & \\
        \midrule

        \multirow[c]{5}{*}{Real}
        & Move Bowl   & 90 & 50 & 45 & 40 & 40 & 53 \\
        & Fold Towel  & 65 & 30 & 25 & 20 & 20 & 32 \\
        & Place Mug   & 75 & 35 & 35 & 35 & 35 & 43 \\
        & Open Drawer & 60 & 35 & 30 & 25 & 25 & 35 \\
        \cmidrule(lr){2-8}
        & \textit{Mean} & 72.5 & 37.5 & 33.8 & 30.0 & 30.0 & 40.8 \\
        \midrule

        \multirow[c]{5}{*}{%
            {\renewcommand{\arraystretch}{1.0}
            \begin{tabular}{@{}r@{\hspace{3pt}}l@{}}
                \multicolumn{2}{c}{Real} \\
                $+$ & Spatial-Gen
            \end{tabular}}}
        & Move Bowl   & 90 & 75 & 50 & 55 & 45 & 63 \\
        & Fold Towel  & 70 & 55 & 30 & 30 & 25 & 42 \\
        & Place Mug   & 80 & 65 & 40 & 45 & 40 & 54 \\
        & Open Drawer & 65 & 55 & 35 & 40 & 30 & 45 \\
        \cmidrule(lr){2-8}
        & \textit{Mean} & \secondcell{76.3} & \secondcell{62.5} & \secondcell{38.8} & \secondcell{42.5} & \secondcell{35.0} & \secondcell{51.0} \\
        \midrule

        \multirow[c]{5}{*}{%
            {\renewcommand{\arraystretch}{1.0}
            \begin{tabular}{@{}r@{\hspace{3pt}}l@{}}
                \multicolumn{2}{c}{Real} \\
                $+$ & Spatial-Gen \\
                $+$ & Diverse-Gen
            \end{tabular}}}
        & Move Bowl   & 100 & 80 & 85 & 85 & 65 & 83 \\
        & Fold Towel  & 65  & 65 & 65 & 60 & 40 & 59 \\
        & Place Mug   & 90  & 70 & 75 & 75 & 55 & 73 \\
        & Open Drawer & 65  & 65 & 65 & 60 & 50 & 61 \\
        \cmidrule(lr){2-8}
        & \textit{Mean} & \bestcell{80.0} & \bestcell{70.0} & \bestcell{72.5} & \bestcell{70.0} & \bestcell{52.5} & \bestcell{69.0} \\
        \bottomrule
    \end{tabular*}
\end{table*}

\paragraph{Evaluation settings}
We evaluate each policy under one in-distribution setting and four out-of-distribution settings. In the \emph{Standard setup}, the scene, object instances, and lighting remain consistent with the data-collection environment, while object poses are sampled within the training range. \emph{Spatial rearrangement} retains the original objects and scene appearance but expands the range of task-relevant object placements. \emph{Distractors} introduces additional task-irrelevant objects on the tabletop. \emph{Novel instances} replaces the target object with an unseen instance from the same functional category. \emph{Lighting shift} changes the ambient illumination while preserving the task configuration.

\paragraph{Trial protocol and success criteria}
All policy variants are evaluated under identical test conditions. We conduct 20 physical trials for each task--setting pair, resulting in $4\times5\times20=400$ trials per training regime. The environment is reset after each trial, and task success is recorded as a binary outcome. Table~\ref{tab:task_completion_criteria} summarizes the execution stages and completion criteria, where Progress indicates the average normalized temporal position of each stage over successful demonstrations. A trial is considered successful only when the final \textit{Done} criterion is satisfied.

\begin{table*}[t]
    \centering
    \caption{Stage-wise execution protocol for the four real-world manipulation tasks. Progress indicates the average normalized temporal position at which each stage is completed over successful demonstrations, rather than its success rate.}
    \label{tab:task_completion_criteria}
    \small
    \setlength{\tabcolsep}{4pt}
    \renewcommand{\arraystretch}{1.12}
    \begin{tabularx}{\textwidth}{
        @{}
        >{\raggedright\arraybackslash}p{0.12\textwidth}
        >{\raggedright\arraybackslash}p{0.20\textwidth}
        >{\centering\arraybackslash}p{0.08\textwidth}
        >{\raggedright\arraybackslash}X
        @{}
    }
        \toprule
        \textbf{Task}
        & \textbf{Stage}
        & \textbf{Progress}
        & \textbf{Completion criterion} \\
        \midrule

        \multirow[t]{2}{*}{\textbf{Move Bowl}}
        & S1: Approach and grasp
        & 45\%
        & The robot approaches the target bowl and establishes a stable grasp. \\
        & S2: Lift and Transport
        & 87\%
        & The bowl is picked up and moved toward the target. \\
        & \textit{Done}: Place on the target
        & 100\%
        & The bowl is placed at the designated target position and remains stable. \\
        \cmidrule(lr){1-4}

        \multirow[t]{4}{*}{\textbf{Fold Towel}}
        & S1: Approach and grasp
        & 33\%
        & The left and right grippers approach the towel and grasp the upper side required for the first fold. \\
        & S2: Fold top to bottom
        & 68\%
        & The upper part of the towel is folded downward over the lower part. \\
        & S3: Regrasp the right side
        & 77\%
        & The robot repositions its grasp on the right side to prepare for the second fold. \\
        & \textit{Done}: Fold right to left
        & 100\%
        & The right part is folded toward the left, producing the specified two-stage folded configuration. \\
        \cmidrule(lr){1-4}

        \multirow[t]{2}{*}{\textbf{Place Mug}}
        & S1: Approach and grasp
        & 49\%
        & The robot approaches the mug, aligns the gripper with its handle, and establishes a stable grasp on the handle. \\
        & S2: Lift and Transport
        & 89\%
        & The mug is picked up and moved toward the target. \\
        & \textit{Done}: Move to the target
        & 100\%
        & The mug is moved to the designated target location and remains stably placed within the target region. \\
        \cmidrule(lr){1-4}

        \multirow[t]{4}{*}{\textbf{Open Drawer}}
        & S1: Grasp the handle
        & 20\%
        & The robot's right arm grasps the drawer handle. \\
        & S2: Pull the drawer
        & 38\%
        & The robot opens the drawer sufficiently for object placement. \\
        & S3: Grasp the marker
        & 55\%
        & The left robot arm approaches the marker and establishes a stable grasp. \\
        & S4: Place the marker
        & 76\%
        & The marker is transported into the open drawer and released inside it. \\
        & \textit{Done}: Close the drawer
        & 100\%
        & The drawer is fully closed with the marker remaining inside, completing the long-horizon task. \\
        \bottomrule
    \end{tabularx}
\end{table*}

\begin{figure*}[!t]
    \centering
    \includegraphics[width=0.96\textwidth]{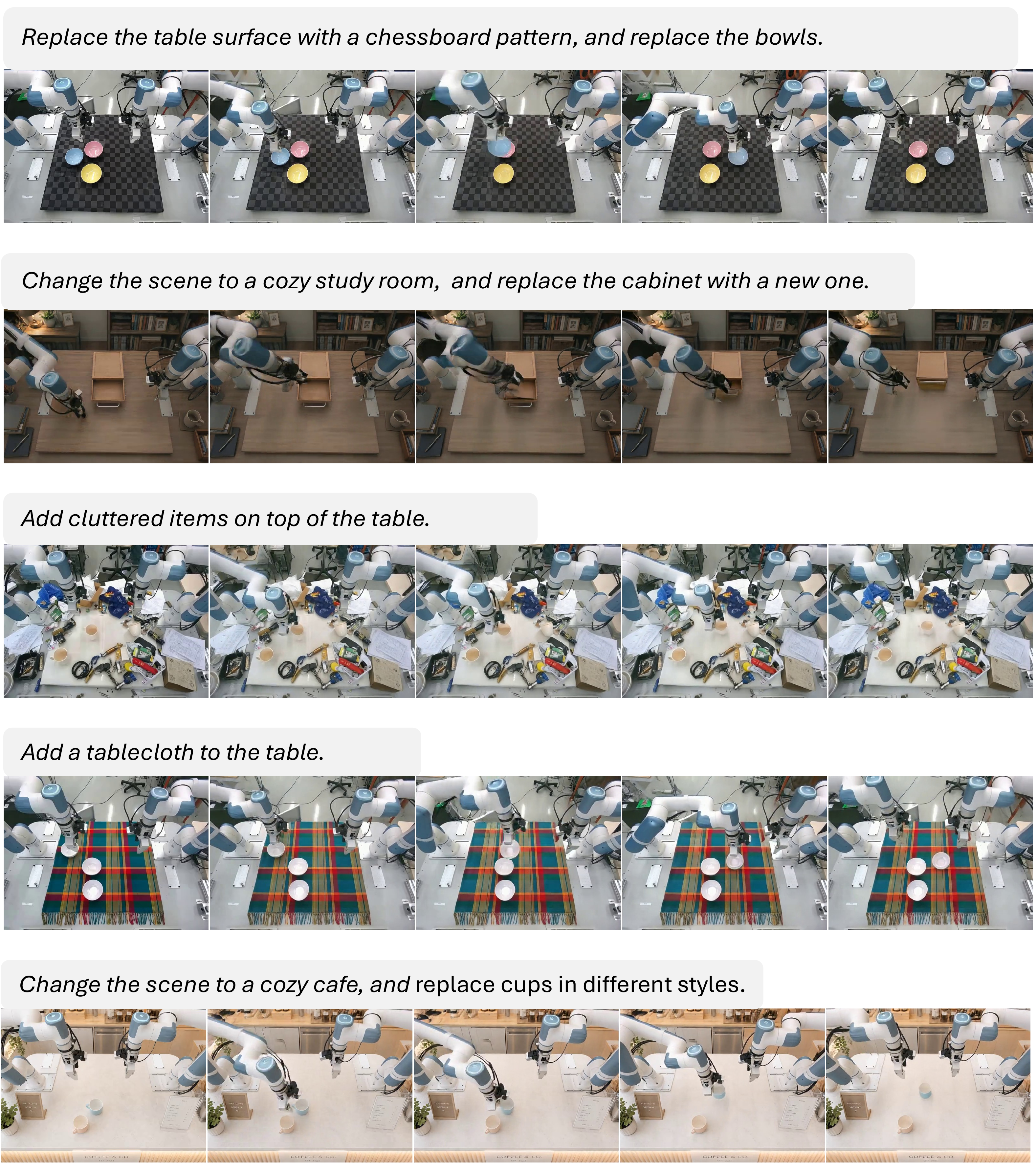}
    % \vspace{-2em}
    \caption{\textbf{Real-world video rollouts under out-of-domain (OOD) conditions.} GeniWorld maintains precise motion trajectories and realistic physical interactions even when exposed to unseen object instances, novel backgrounds, and complex environmental distractors. }
    \vspace{-0.8em}
    \label{fig:demo_realworld_ood}
\end{figure*}

\begin{figure*}[!t]
    \centering
    \includegraphics[width=0.95\textwidth]{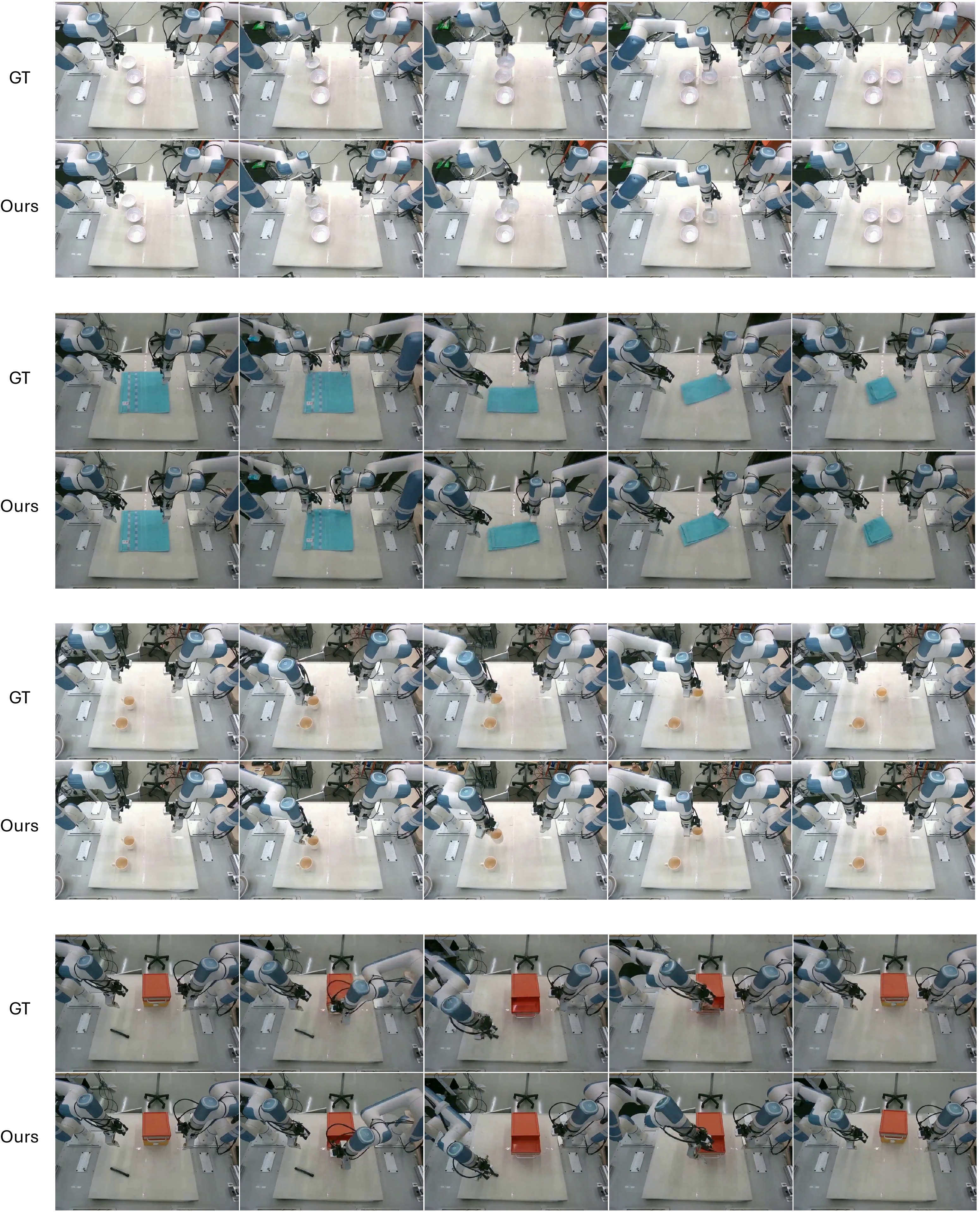}
    % \vspace{-2em}
    \caption{\textbf{Qualitative results of GeniWorld on in-domain real-world tasks.}  The predicted video sequences demonstrate accurate spatial trajectories, fine-grained object interaction, and strict visual temporal consistency aligned with ground-truth rollouts. }
    \vspace{-0.8em}
    \label{fig:demo_realworld_id}
\end{figure*}

\begin{figure*}[!t]
    \centering
    \includegraphics[width=0.95\textwidth]{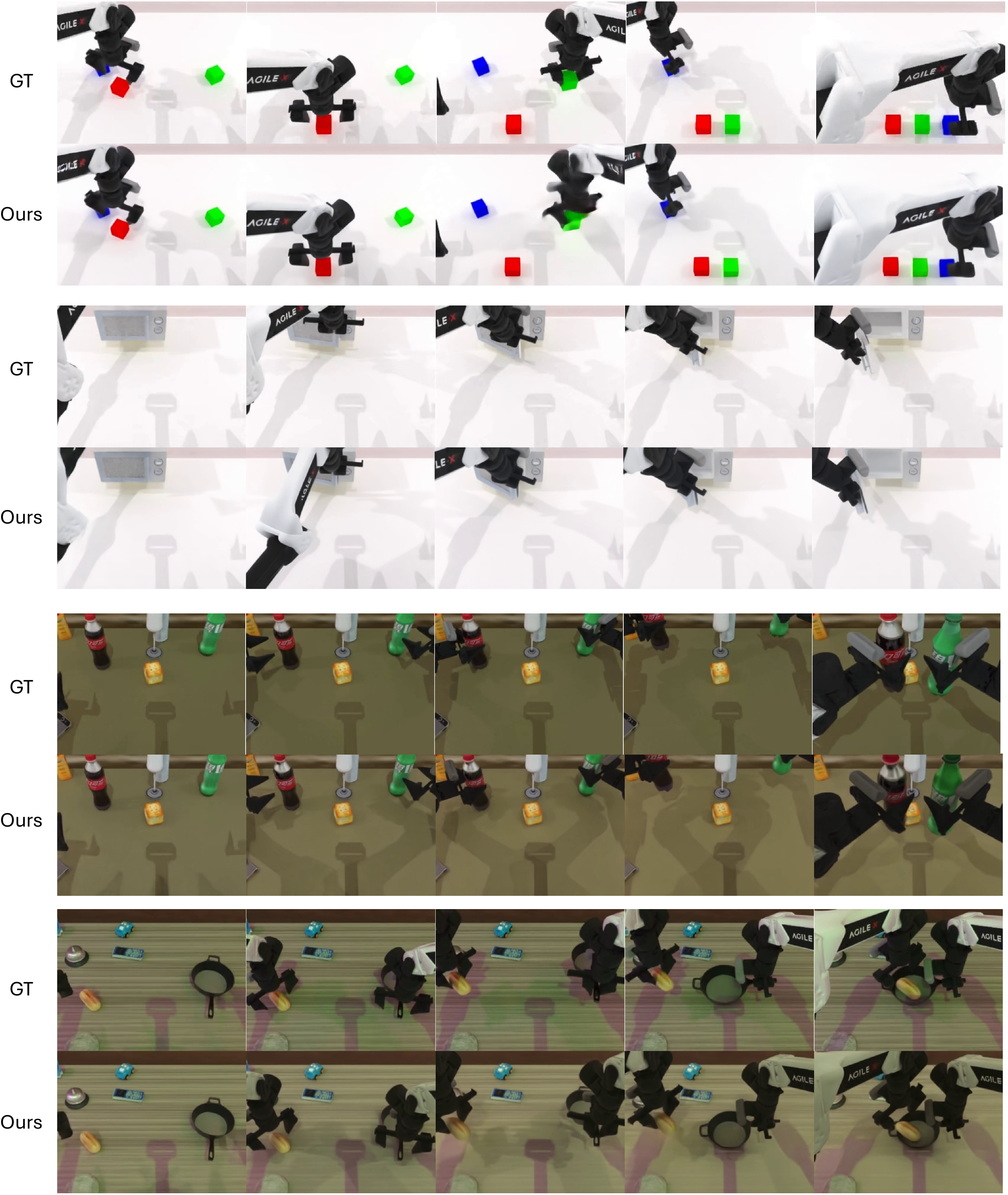}
    % \vspace{-2em}
    \caption{\textbf{Qualitative visualizations of generated rollouts on RoboTwin.}  We show the predictions across Clean-to-Clean (in-domain) and Clean-to-Random (out-of-distribution) setups. GeniWorld maintains high visual fidelity and precise temporal consistency even under randomized environmental perturbations.}
    \vspace{-0.8em}
    \label{fig:demo_robotwin}
\end{figure*}

\begin{figure*}[!t]
    \centering
    \includegraphics[width=0.95\textwidth]{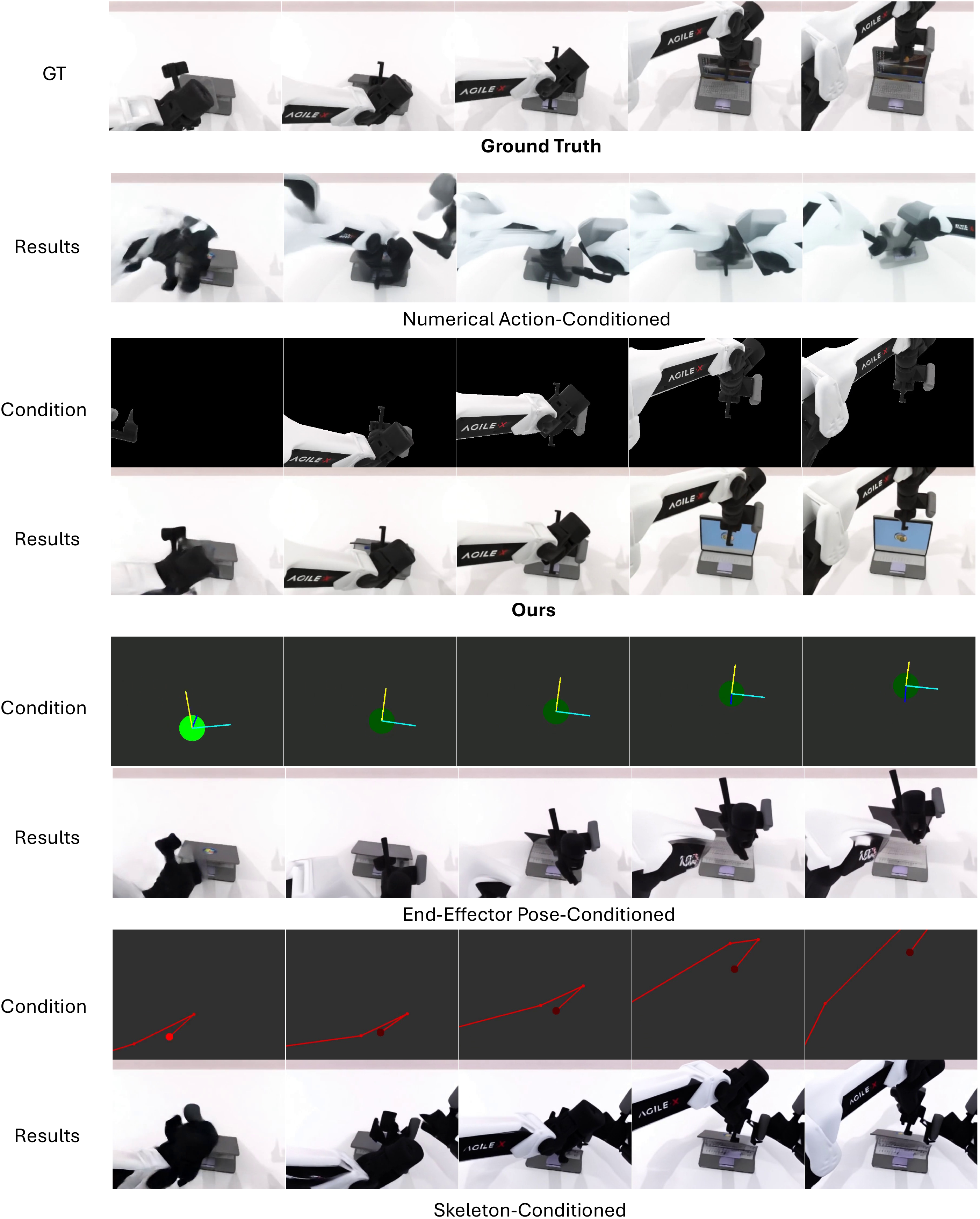}
    % \vspace{-2em}
    \caption{\textbf{Qualitative comparison of generation rollouts across different action representations.}  }
    \vspace{-0.8em}
    \label{fig:demo_robotwin_ac}
\end{figure*}

%% file: references.bib
@article{wan2025wan,
  title   = {Wan: Open and Advanced Large-Scale Video Generative Models},
  author  = {{Wan Team}},
  journal = {arXiv preprint arXiv:2503.20314},
  year    = {2025}
}

@inproceedings{zhu2025irasim,
  title     = {{IRASim}: A Fine-Grained World Model for Robot Manipulation},
  author    = {Zhu, Fangqi and Wu, Hongtao and Guo, Song and Liu, Yuxuan and Cheang, Chi and Kong, Tao},
  booktitle = {Proceedings of the IEEE/CVF International Conference on Computer Vision},
  year      = {2025}
}

@article{jiang2025enerverseac,
  title   = {{EnerVerse-AC}: Envisioning Embodied Environments with Action Condition},
  author  = {Jiang, Yutong and Chen, Siyi and Huang, Siyuan and Chen, Liliang and Zhou, Pengfei and Liao, Yue and others},
  journal = {arXiv preprint arXiv:2505.09723},
  year    = {2025}
}

@article{guo2025ctrlworld,
  title   = {{Ctrl-World}: A Controllable Generative World Model for Robot Manipulation},
  author  = {Guo, Yanjiang and Shi, Lucy Xiaoyang and Chen, Jiaben and Finn, Chelsea},
  journal = {arXiv preprint arXiv:2510.10125},
  year    = {2025}
}

@inproceedings{wang2025vap,
  title     = {Precise Action-to-Video Generation through Visual Action Prompts},
  author    = {Wang, Yiming and Wen, Chuanxia and Guo, Haoyu and Peng, Songyou and Qin, Ming and Bao, Hujun and others},
  booktitle = {Proceedings of the IEEE/CVF International Conference on Computer Vision},
  pages     = {12713--12724},
  year      = {2025}
}

@article{wu2026oscar,
  title   = {{OSCAR}: Omni-Embodiment Action-Conditioned World Model for Robotics},
  author  = {Wu, Zhen and Gao, Jun},
  journal = {arXiv preprint arXiv:2606.04463},
  year    = {2026}
}

@article{chen2026bridgev2w,
  title   = {{BridgeV2W}: Bridging Video Generation Models to Embodied World Models via Embodiment Masks},
  author  = {Chen, Yiming and Li, Peiyan and Yang, Jian and He, Kai and Wu, Xinyu and Xu, Yifan and others},
  journal = {arXiv preprint arXiv:2602.03793},
  year    = {2026}
}

@article{li2025worldeval,
  title   = {{WorldEval}: World Model as Real-World Robot Policies Evaluator},
  author  = {Li, Yichen and Zhu, Yifeng and Wen, Jian and Shen, Chunhua and Xu, Yang},
  journal = {arXiv preprint arXiv:2505.19017},
  year    = {2025}
}

@article{quevedo2025worldgym,
  title   = {{WorldGym}: World Model as an Environment for Policy Evaluation},
  author  = {Quevedo, Javier and Sharma, Aman Kumar and Sun, Yanchao and Suryavanshi, Vedant and Liang, Percy and Yang, Sherry},
  journal = {arXiv preprint arXiv:2506.00613},
  year    = {2025}
}

@article{shang2026worldarena,
  title   = {{WorldArena}: A Unified Benchmark for Evaluating Perception and Functional Utility of Embodied World Models},
  author  = {Shang, Yu and Li, Zhuohang and Ma, Yiding and Su, Weikang and Jin, Xin and Wang, Ziyou and others},
  journal = {arXiv preprint arXiv:2602.08971},
  year    = {2026}
}

@article{gao2026dreamdojo,
  title   = {{DreamDojo}: A Generalist Robot World Model from Large-Scale Human Videos},
  author  = {Gao, Shenyuan and Liang, Wenqi and Zheng, Kaiyuan and Malik, Aqib and Ye, Seonghyeon and Yu, Sihyun and others},
  journal = {arXiv preprint arXiv:2602.06949},
  year    = {2026}
}

@misc{openai2025gptimage,
  author       = {{OpenAI}},
  title        = {Introducing Our Latest Image Generation Model in the {API}},
  year         = {2025},
  month        = apr,
  howpublished = {\url{https://openai.com/index/image-generation-api/}}
}

@online{251215840LargeVideo,
  title = {[2512.15840] {{Large Video Planner Enables Generalizable Robot Control}}},
  url = {https://arxiv.org/abs/2512.15840},
  urldate = {2026-08-01}
}

@article{goodmanWorldModelsDavid,
  title = {World {{Models}} - {{David Ha}}, {{Jürgen Schmidhuber}}},
  author = {Goodman, Edmund}
}

@online{hafnerDreamControlLearning2020,
  title = {Dream to {{Control}}: {{Learning Behaviors}} by {{Latent Imagination}}},
  author = {Hafner, Danijar and Lillicrap, Timothy and Ba, Jimmy and Norouzi, Mohammad},
  date = {2020-03-17},
  eprint = {1912.01603},
  eprinttype = {arXiv},
  eprintclass = {cs.LG},
  doi = {10.48550/arXiv.1912.01603},
  pubstate = {prepublished}
}

@online{houWorldModelRobot2026,
  title = {World {{Model}} for {{Robot Learning}}: {{A Comprehensive Survey}}},
  author = {Hou, Bohan and Li, Gen and Jia, Jindou and An, Tuo and Guo, Xinying and Leng, Sicong and Geng, Haoran and Ze, Yanjie and Harada, Tatsuya and Torr, Philip and Mees, Oier and Pollefeys, Marc and Liu, Zhuang and Wu, Jiajun and Abbeel, Pieter and Malik, Jitendra and Du, Yilun and Yang, Jianfei},
  date = {2026-04-30},
  eprint = {2605.00080},
  eprinttype = {arXiv},
  eprintclass = {cs.RO},
  doi = {10.48550/arXiv.2605.00080},
  pubstate = {prepublished}
}

@article{wu2024ivideogpt,
  title={ivideogpt: Interactive videogpts are scalable world models},
  author={Wu, Jialong and Yin, Shaofeng and Feng, Ningya and He, Xu and Li, Dong and Hao, Jianye and Long, Mingsheng},
  journal={Advances in Neural Information Processing Systems},
  volume={37},
  pages={68082--68119},
  year={2024}
}

@online{nvidiaCosmosWorldFoundation2025,
  title = {Cosmos {{World Foundation Model Platform}} for {{Physical AI}}},
  author = {NVIDIA and Agarwal, Niket and Ali, Arslan and Bala, Maciej and Balaji, Yogesh and Barker, Erik and Cai, Tiffany and Chattopadhyay, Prithvijit and Chen, Yongxin and Cui, Yin and Ding, Yifan and Dworakowski, Daniel and Fan, Jiaojiao and Fenzi, Michele and Ferroni, Francesco and Fidler, Sanja and Fox, Dieter and Ge, Songwei and Ge, Yunhao and Gu, Jinwei and Gururani, Siddharth and He, Ethan and Huang, Jiahui and Huffman, Jacob and Jannaty, Pooya and Jin, Jingyi and Kim, Seung Wook and Klár, Gergely and Lam, Grace and Lan, Shiyi and Leal-Taixe, Laura and Li, Anqi and Li, Zhaoshuo and Lin, Chen-Hsuan and Lin, Tsung-Yi and Ling, Huan and Liu, Ming-Yu and Liu, Xian and Luo, Alice and Ma, Qianli and Mao, Hanzi and Mo, Kaichun and Mousavian, Arsalan and Nah, Seungjun and Niverty, Sriharsha and Page, David and Paschalidou, Despoina and Patel, Zeeshan and Pavao, Lindsey and Ramezanali, Morteza and Reda, Fitsum and Ren, Xiaowei and Sabavat, Vasanth Rao Naik and Schmerling, Ed and Shi, Stella and Stefaniak, Bartosz and Tang, Shitao and Tchapmi, Lyne and Tredak, Przemek and Tseng, Wei-Cheng and Varghese, Jibin and Wang, Hao and Wang, Haoxiang and Wang, Heng and Wang, Ting-Chun and Wei, Fangyin and Wei, Xinyue and Wu, Jay Zhangjie and Xu, Jiashu and Yang, Wei and Yen-Chen, Lin and Zeng, Xiaohui and Zeng, Yu and Zhang, Jing and Zhang, Qinsheng and Zhang, Yuxuan and Zhao, Qingqing and Zolkowski, Artur},
  date = {2025-07-09},
  eprint = {2501.03575},
  eprinttype = {arXiv},
  eprintclass = {cs.CV},
  doi = {10.48550/arXiv.2501.03575},
  pubstate = {prepublished}
}

@online{teamEvaluatingGeminiRobotics2026,
  title = {Evaluating {{Gemini Robotics Policies}} in a {{Veo World Simulator}}},
  author = {Team, Gemini Robotics and Choromanski, Krzysztof and Devin, Coline and Du, Yilun and Dwibedi, Debidatta and Gao, Ruiqi and Jindal, Abhishek and Kipf, Thomas and Kirmani, Sean and Leal, Isabel and Liu, Fangchen and Majumdar, Anirudha and Marmon, Andrew and Parada, Carolina and Rubanova, Yulia and Shah, Dhruv and Sindhwani, Vikas and Tan, Jie and Xia, Fei and Xiao, Ted and Yang, Sherry and Yu, Wenhao and Zhou, Allan},
  date = {2026-01-06},
  eprint = {2512.10675},
  eprinttype = {arXiv},
  eprintclass = {cs.RO},
  doi = {10.48550/arXiv.2512.10675},
  pubstate = {prepublished}
}

@online{zhuUnifiedWorldModels2025a,
  title = {Unified {{World Models}}: {{Coupling Video}} and {{Action Diffusion}} for {{Pretraining}} on {{Large Robotic Datasets}}},
  author = {Zhu, Chuning and Yu, Raymond and Feng, Siyuan and Burchfiel, Benjamin and Shah, Paarth and Gupta, Abhishek},
  date = {2025-05-23},
  eprint = {2504.02792},
  eprinttype = {arXiv},
  eprintclass = {cs.RO},
  doi = {10.48550/arXiv.2504.02792},
  pubstate = {prepublished}
}

@inproceedings{bruce2024genie,
  title={Genie: Generative interactive environments},
  author={Bruce, Jake and Dennis, Michael D and Edwards, Ashley and Parker-Holder, Jack and Shi, Yuge and Hughes, Edward and Lai, Matthew and Mavalankar, Aditi and Steigerwald, Richie and Apps, Chris and others},
  booktitle={Forty-first International Conference on Machine Learning},
  year={2024}
}

@article{team2025evaluating,
  title={Evaluating Gemini Robotics Policies in a Veo World Simulator},
  author={Team, Gemini Robotics and Choromanski, Krzysztof and Devin, Coline and Du, Yilun and Dwibedi, Debidatta and Gao, Ruiqi and Jindal, Abhishek and Kipf, Thomas and Kirmani, Sean and Leal, Isabel and others},
  journal={arXiv preprint arXiv:2512.10675},
  year={2025}
}

@article{yang2023learning,
  title={Learning interactive real-world simulators},
  author={Yang, Sherry and Du, Yilun and Ghasemipour, Kamyar and Tompson, Jonathan and Kaelbling, Leslie and Schuurmans, Dale and Abbeel, Pieter},
  journal={arXiv preprint arXiv:2310.06114},
  year={2023}
}

@article{wang2026interactive,
  title={Interactive world simulator for robot policy training and evaluation},
  author={Wang, Yixuan and Syed, Rhythm and Wu, Fangyu and Zhang, Mengchao and Onol, Aykut and Barreiros, Jose and Nayyeri, Hooshang and Dear, Tony and Zhang, Huan and Li, Yunzhu},
  journal={arXiv preprint arXiv:2603.08546},
  year={2026}
}

@article{xu2026kinema4d,
  title={Kinema4d: Kinematic 4d world modeling for spatiotemporal embodied simulation},
  author={Xu, Mutian and Zhang, Tianbao and Liu, Tianqi and Chen, Zhaoxi and Han, Xiaoguang and Liu, Ziwei},
  journal={arXiv preprint arXiv:2603.16669},
  year={2026}
}

@article{qiu2026ge,
  title={Ge-sim 2.0: A roadmap towards comprehensive closed-loop video world simulators for robotic manipulation},
  author={Qiu, Boxiang and Chen, Liliang and Liao, Yue and Wang, Nan and Wang, Lintao and Luo, Jiayi and Zhao, Wenzhi and Chen, Shengcong and Chen, Di and Li, Ye and others},
  journal={arXiv preprint arXiv:2605.27491},
  year={2026}
}

@article{jiang2026wovr,
  title={Wovr: World models as reliable simulators for post-training vla policies with rl},
  author={Jiang, Zhennan and Zhou, Shangqing and Jiang, Yutong and Huang, Zefang and Wei, Mingjie and Chen, Yuhui and Zhou, Tianxing and Guo, Zhen and Lin, Hao and Zhang, Quanlu and others},
  journal={arXiv preprint arXiv:2602.13977},
  year={2026}
}

@article{zhu2025wmpo,
  title={Wmpo: World model-based policy optimization for vision-language-action models},
  author={Zhu, Fangqi and Yan, Zhengyang and Hong, Zicong and Shou, Quanxin and Ma, Xiao and Guo, Song},
  journal={arXiv preprint arXiv:2511.09515},
  year={2025}
}

@article{gao2026sword,
  title={Sword: Style-robust world models as simulators via dynamic latent bootstrapping for vla policy post-training},
  author={Gao, Jiaxuan and Guo, Yongjian and Guan, Zhong and Huang, Wen and Ma, Wanlun and Xiao, Xi and Xiong, Junwu and Wen, Sheng},
  journal={arXiv preprint arXiv:2605.07288},
  year={2026}
}

@article{li2024robogsim,
  title={Robogsim: A real2sim2real robotic gaussian splatting simulator},
  author={Li, Xinhai and Li, Jialin and Zhang, Ziheng and Zhang, Rui and Jia, Fan and Wang, Tiancai and Fan, Haoqiang and Tseng, Kuo-Kun and Wang, Ruiping},
  journal={arXiv preprint arXiv:2411.11839},
  year={2024}
}

@article{zhao2026high,
  title={High-fidelity simulated data generation for real-world zero-shot robotic manipulation learning with gaussian splatting},
  author={Zhao, Haoyu and Zeng, Cheng and Zhuang, Linghao and Zhao, Yaxi and Xue, Shengke and Wang, Hao and Zhao, Xingyue and Li, Zhongyu and Li, Kehan and Huang, Siteng and others},
  journal={IEEE Robotics and Automation Letters},
  year={2026},
  publisher={IEEE}
}

@article{yang2025novel,
  title={Novel demonstration generation with gaussian splatting enables robust one-shot manipulation},
  author={Yang, Sizhe and Yu, Wenye and Zeng, Jia and Lv, Jun and Ren, Kerui and Lu, Cewu and Lin, Dahua and Pang, Jiangmiao},
  journal={arXiv preprint arXiv:2504.13175},
  year={2025}
}

@article{gu2025igen,
  title={IGen: Scalable Data Generation for Robot Learning from Open-World Images},
  author={Gu, Chenghao and Kang, Haolan and Lin, Junchao and Wang, Jinghe and Wu, Duo and Xie, Shuzhao and Huang, Fanding and Ge, Junchen and Gong, Ziyang and Li, Letian and others},
  journal={arXiv preprint arXiv:2512.01773},
  year={2025}
}

@article{ye2025anchordream,
  title={AnchorDream: Repurposing Video Diffusion for Embodiment-Aware Robot Data Synthesis},
  author={Ye, Junjie and Xue, Rong and Van Hoorick, Basile and Tokmakov, Pavel and Irshad, Muhammad Zubair and Wang, Yue and Guizilini, Vitor},
  journal={arXiv preprint arXiv:2512.11797},
  year={2025}
}

@article{c,
  title={Scaling robot learning with semantically imagined experience},
  author={Yu, Tianhe and Xiao, Ted and Stone, Austin and Tompson, Jonathan and Brohan, Anthony and Wang, Su and Singh, Jaspiar and Tan, Clayton and Peralta, Jodilyn and Ichter, Brian and others},
  journal={arXiv preprint arXiv:2302.11550},
  year={2023}
}

@inproceedings{yuan2025roboengine,
  title={Roboengine: Plug-and-play robot data augmentation with semantic robot segmentation and background generation},
  author={Yuan, Chengbo and Joshi, Suraj and Zhu, Shaoting and Su, Hang and Zhao, Hang and Gao, Yang},
  booktitle={2025 IEEE/RSJ International Conference on Intelligent Robots and Systems (IROS)},
  pages={7622--7629},
  year={2025},
  organization={IEEE}
}

@article{alhaija2025cosmos,
  title={Cosmos-transfer1: Conditional world generation with adaptive multimodal control},
  author={Alhaija, Hassan Abu and Alvarez, Jose and Bala, Maciej and Cai, Tiffany and Cao, Tianshi and Cha, Liz and Chen, Joshua and Chen, Mike and Ferroni, Francesco and Fidler, Sanja and others},
  journal={arXiv preprint arXiv:2503.14492},
  year={2025}
}

@article{jang2025dreamgen,
  title={Dreamgen: Unlocking generalization in robot learning through video world models},
  author={Jang, Joel and Ye, Seonghyeon and Lin, Zongyu and Xiang, Jiannan and Bjorck, Johan and Fang, Yu and Hu, Fengyuan and Huang, Spencer and Kundalia, Kaushil and Lin, Yen-Chen and others},
  journal={arXiv preprint arXiv:2505.12705},
  year={2025}
}

@article{black2024pi_0,
  title={{\(\pi_0\)}: A Vision-Language-Action Flow Model for General Robot Control},
  author={Black, Kevin and Brown, Noah and Driess, Danny and Esmail, Adnan and Equi, Michael and Finn, Chelsea and Fusai, Niccolo and Groom, Lachy and Hausman, Karol and Ichter, Brian and others},
  journal={arXiv preprint arXiv:2410.24164},
  year={2024}
}

@article{intelligence2025pi_,
  title={{\(\pi_{0.5}\)}: A Vision-Language-Action Model with Open-World Generalization},
  author={Intelligence, Physical and Black, Kevin and Brown, Noah and Darpinian, James and Dhabalia, Karan and Driess, Danny and Esmail, Adnan and Equi, Michael and Finn, Chelsea and Fusai, Niccolo and others},
  journal={arXiv preprint arXiv:2504.16054},
  year={2025}
}

@article{intelligence2025pi,
  title={{\(\pi^\ast_{0.6}\)}: A VLA That Learns From Experience},
  author={Intelligence, Physical and Amin, Ali and Aniceto, Raichelle and Balakrishna, Ashwin and Black, Kevin and Conley, Ken and Connors, Grace and Darpinian, James and Dhabalia, Karan and DiCarlo, Jared and others},
  journal={arXiv preprint arXiv:2511.14759},
  year={2025}
}

@article{intelligence2026pi,
  title={{\(\pi_{0.7}\)}: A Steerable Generalist Robotic Foundation Model with Emergent Capabilities},
  author={Intelligence, Physical and Ai, Bo and Amin, Ali and Aniceto, Raichelle and Balakrishna, Ashwin and Balke, Greg and Black, Kevin and Bokinsky, George and Cao, Shihao and Charbonnier, Thomas and others},
  journal={arXiv preprint arXiv:2604.15483},
  year={2026}
}

@article{bjorck2025gr00t,
  title={Gr00t n1: An open foundation model for generalist humanoid robots},
  author={Bjorck, Johan and Casta{\~n}eda, Fernando and Cherniadev, Nikita and Da, Xingye and Ding, Runyu and Fan, Linxi and Fang, Yu and Fox, Dieter and Hu, Fengyuan and Huang, Spencer and others},
  journal={arXiv preprint arXiv:2503.14734},
  year={2025}
}

@article{team2025gemini,
  title={Gemini robotics: Bringing ai into the physical world},
  author={Team, Gemini Robotics and Abeyruwan, Saminda and Ainslie, Joshua and Alayrac, Jean-Baptiste and Arenas, Montserrat Gonzalez and Armstrong, Travis and Balakrishna, Ashwin and Baruch, Robert and Bauza, Maria and Blokzijl, Michiel and others},
  journal={arXiv preprint arXiv:2503.20020},
  year={2025}
}

@article{lin2026posevla,
  title={PoseVLA: Universal Pose Pretraining for Generalizable Vision-Language-Action Policies},
  author={Lin, Haitao and Yu, Hanyang and Huang, Jingshun and Zhang, He and Ling, Yonggen and Tan, Ping and Xue, Xiangyang and Fu, Yanwei},
  journal={arXiv preprint arXiv:2602.19710},
  year={2026}
}

@article{li2026causal,
  title={Causal world modeling for robot control},
  author={Li, Lin and Zhang, Qihang and Luo, Yiming and Yang, Shuai and Wang, Ruilin and Han, Fei and Yu, Mingrui and Gao, Zelin and Xue, Nan and Zhu, Xing and others},
  journal={arXiv preprint arXiv:2601.21998},
  year={2026}
}

@article{ye2026world,
  title={World action models are zero-shot policies},
  author={Ye, Seonghyeon and Ge, Yunhao and Zheng, Kaiyuan and Gao, Shenyuan and Yu, Sihyun and Kurian, George and Indupuru, Suneel and Tan, You Liang and Zhu, Chuning and Xiang, Jiannan and others},
  journal={arXiv preprint arXiv:2602.15922},
  year={2026}
}

@article{yuan2026fast,
  title={Fast-wam: Do world action models need test-time future imagination?},
  author={Yuan, Tianyuan and Dong, Zibin and Liu, Yicheng and Zhao, Hang},
  journal={arXiv preprint arXiv:2603.16666},
  year={2026}
}

@article{zhou2025libero,
  title={Libero-pro: Towards robust and fair evaluation of vision-language-action models beyond memorization},
  author={Zhou, Xueyang and Xu, Yangming and Tie, Guiyao and Chen, Yongchao and Zhang, Guowen and Chu, Duanfeng and Zhou, Pan and Sun, Lichao},
  journal={arXiv preprint arXiv:2510.03827},
  year={2025}
}

@article{fei2025libero,
  title={Libero-plus: In-depth robustness analysis of vision-language-action models},
  author={Fei, Senyu and Wang, Siyin and Shi, Junhao and Dai, Zihao and Cai, Jikun and Qian, Pengfang and Ji, Li and He, Xinzhe and Zhang, Shiduo and Fei, Zhaoye and others},
  journal={arXiv preprint arXiv:2510.13626},
  year={2025}
}

@article{chen2025robotwin,
  title={Robotwin 2.0: A scalable data generator and benchmark with strong domain randomization for robust bimanual robotic manipulation},
  author={Chen, Tianxing and Chen, Zanxin and Chen, Baijun and Cai, Zijian and Liu, Yibin and Li, Zixuan and Liang, Qiwei and Lin, Xianliang and Ge, Yiheng and Gu, Zhenyu and others},
  journal={arXiv preprint arXiv:2506.18088},
  year={2025}
}

@article{yuan2026qwen,
  title={Qwen-robotmanip technical report: Alignment unlocks scale for robotic manipulation foundation models},
  author={Yuan, Haoqi and Liang, Zhixuan and Chen, Anzhe and Wang, Ye and Li, Haoyang and Lin, Pei and Huang, Yiyang and Lei, Zixing and Zhang, Tong and Zhang, Jiazhao and others},
  journal={arXiv preprint arXiv:2606.17846},
  year={2026}
}

@article{yu2026maskwam,
  title={Maskwam: Unifying mask prompting and prediction for world-action models},
  author={Yu, Hanyang and Lin, Haitao and Zhang, Jingbo and Zhang, Wenyao and Gu, Chenghao and Li, Heng and Tan, Ping},
  journal={arXiv preprint arXiv:2606.13515},
  year={2026}
}

@article{assran2025v,
  title={V-jepa 2: Self-supervised video models enable understanding, prediction and planning},
  author={Assran, Mido and Bardes, Adrien and Fan, David and Garrido, Quentin and Howes, Russell and Muckley, Matthew and Rizvi, Ammar and Roberts, Claire and Sinha, Koustuv and Zholus, Artem and others},
  journal={arXiv preprint arXiv:2506.09985},
  year={2025}
}

@article{huang2026self,
  title={Self forcing: Bridging the train-test gap in autoregressive video diffusion},
  author={Huang, Xun and Li, Zhengqi and He, Guande and Zhou, Mingyuan and Shechtman, Eli},
  journal={Advances in Neural Information Processing Systems},
  volume={38},
  pages={167283--167308},
  year={2026}
}

@article{lipman2022flow,
  title={Flow matching for generative modeling},
  author={Lipman, Yaron and Chen, Ricky TQ and Ben-Hamu, Heli and Nickel, Maximilian and Le, Matt},
  journal={arXiv preprint arXiv:2210.02747},
  year={2022}
}

@article{wang2004image,
  title={Image quality assessment: from error visibility to structural similarity},
  author={Wang, Zhou and Bovik, Alan C and Sheikh, Hamid R and Simoncelli, Eero P},
  journal={IEEE transactions on image processing},
  volume={13},
  number={4},
  pages={600--612},
  year={2004},
  publisher={IEEE}
}

@inproceedings{zhang2018unreasonable,
  title={The unreasonable effectiveness of deep features as a perceptual metric},
  author={Zhang, Richard and Isola, Phillip and Efros, Alexei A and Shechtman, Eli and Wang, Oliver},
  booktitle={2018 IEEE/CVF conference on computer vision and pattern recognition},
  pages={586--595},
  year={2018},
  organization={IEEE}
}

@article{heusel2017gans,
  title={Gans trained by a two time-scale update rule converge to a local nash equilibrium},
  author={Heusel, Martin and Ramsauer, Hubert and Unterthiner, Thomas and Nessler, Bernhard and Hochreiter, Sepp},
  journal={Advances in neural information processing systems},
  volume={30},
  year={2017}
}

@article{unterthiner2018towards,
  title={Towards accurate generative models of video: A new metric \& challenges},
  author={Unterthiner, Thomas and Van Steenkiste, Sjoerd and Kurach, Karol and Marinier, Raphael and Michalski, Marcin and Gelly, Sylvain},
  journal={arXiv preprint arXiv:1812.01717},
  year={2018}
}

@article{bai2025qwen3,
  title={Qwen3-vl technical report},
  author={Bai, Shuai and Cai, Yuxuan and Chen, Ruizhe and Chen, Keqin and Chen, Xionghui and Cheng, Zesen and Deng, Lianghao and Ding, Wei and Gao, Chang and Ge, Chunjiang and others},
  journal={arXiv preprint arXiv:2511.21631},
  year={2025}
}

@article{achiam2023gpt,
  title={Gpt-4 technical report},
  author={Achiam, Josh and Adler, Steven and Agarwal, Sandhini and Ahmad, Lama and Akkaya, Ilge and Aleman, Florencia Leoni and Almeida, Diogo and Altenschmidt, Janko and Altman, Sam and Anadkat, Shyamal and others},
  journal={arXiv preprint arXiv:2303.08774},
  year={2023}
}

@article{wu2025qwen,
  title={Qwen-image technical report},
  author={Wu, Chenfei and Li, Jiahao and Zhou, Jingren and Lin, Junyang and Gao, Kaiyuan and Yan, Kun and Yin, Sheng-ming and Bai, Shuai and Xu, Xiao and Chen, Yilei and others},
  journal={arXiv preprint arXiv:2508.02324},
  year={2025}
}

@article{zhang2026dreamvla,
  title={Dreamvla: a vision-language-action model dreamed with comprehensive world knowledge},
  author={Zhang, Wenyao and Liu, Hongsi and Qi, Zekun and Wang, Yunnan and Yu, Xinqiang and Zhang, Jiazhao and Dong, Runpei and He, Jiawei and Wang, He and Zhang, Zhizheng and others},
  journal={Advances in Neural Information Processing Systems},
  volume={38},
  pages={24195--24228},
  year={2026}
}

@article{zhang2026disentangled,
  title={Disentangled robot learning via separate forward and inverse dynamics pretraining},
  author={Zhang, Wenyao and Zhang, Bozhou and Qi, Zekun and Zeng, Wenjun and Jin, Xin and Zhang, Li},
  journal={arXiv preprint arXiv:2604.16391},
  year={2026}
}

@inproceedings{zhang2023adding,
  title={Adding conditional control to text-to-image diffusion models},
  author={Zhang, Lvmin and Rao, Anyi and Agrawala, Maneesh},
  booktitle={2023 IEEE/CVF International Conference on Computer Vision (ICCV)},
  pages={3813--3824},
  year={2023},
  organization={IEEE}
}

@article{chu2026wan,
  title={Wan-move: Motion-controllable video generation via latent trajectory guidance},
  author={Chu, Ruihang and He, Yefei and Chen, Zhekai and Zhang, Shiwei and Xu, Xiaogang and WANG, Dingdong and Yi, Hongwei and Liu, Xihui and Zhao, Hengshuang and Liu, Yu and others},
  journal={Advances in Neural Information Processing Systems},
  volume={38},
  pages={404--432},
  year={2026}
}

@article{bai2026ioi,
  title={IOI: Decoupling Kinematics and Physics for Interactive World Models},
  author={Bai, Chengyu and Jia, Peidong and Guo, Tiecheng and Wang, Yukai and Ma, Rui and Zhao, Fangyuan and Fan, Chunkai and Wei, Xiaobao and Chen, Jintao and Wang, Hao and others},
  journal={arXiv preprint arXiv:2606.23296},
  year={2026}
}

@article{alzayer2026masked,
  title={Masked Visual Actions for Unified World Modeling},
  author={Alzayer, Hadi and Huang, Wenlong and Chen, Haonan and Luey, Christopher and Zhang, Lvmin and Agrawala, Maneesh and Wetzstein, Gordon and Fei-Fei, Li and Du, Yilun and Wu, Jiajun and others},
  journal={arXiv preprint arXiv:2607.19343},
  year={2026}
}

@article{kim2026robot,
  title={Robot-Factored World Models via Robot Rendering},
  author={Kim, Byungjun and Kim, Taeksoo and Cha, Hyunsoo and Joo, Hanbyul},
  journal={arXiv preprint arXiv:2607.22535},
  year={2026}
}

@article{yang2026ea,
  title={EA-WM: Event-Aware Generative World Model with Structured Kinematic-to-Visual Action Fields},
  author={Yang, Zhaoyang and Jin, Yurun and Qi, Lizhe and Huang, Cong and Chen, Kai},
  journal={arXiv preprint arXiv:2605.06192},
  year={2026}
}
